\documentclass[a4paper,conference]{IEEEtran}

\usepackage{placeins}
\usepackage{graphicx}
\usepackage{subfigure}	
\usepackage{makecell}
\usepackage{float}
\usepackage{pifont}
\usepackage{soul}
\newcommand{\cmark}{\ding{51}}
\newcommand{\xmark}{\ding{55}}
\usepackage{xspace}
\makeatletter
\DeclareRobustCommand\onedot{\futurelet\@let@token\@onedot}
\def\@onedot{\ifx\@let@token.\else.\null\fi\xspace}

\def\eg{\emph{e.g}\onedot} 
\def\ie{\emph{i.e}\onedot}

\def\etal{\emph{et al}\onedot}
\makeatother

\usepackage{booktabs}
\usepackage{amsmath}

\DeclareMathOperator*{\softmax}{softmax}

\usepackage[usenames,dvipsnames]{xcolor}
\newcommand{\red}[1]{\textcolor{red}{#1}}

\newcommand{\green}[1]{\textcolor{OliveGreen}{#1}}

\usepackage{tikz}
\usepackage{pgfplots}
\pgfplotsset{compat=1.14}
\usetikzlibrary{positioning}
\usetikzlibrary{arrows,automata}
\usetikzlibrary{decorations.text}
\usetikzlibrary{patterns}
\usepgfplotslibrary{statistics}

\ifCLASSINFOpdf
\else
\fi
\newcommand{\mypar}[1]{\vspace{0.3cm}\noindent\textbf{#1}}
\usepackage{etoolbox,lipsum}

\usepackage{hyperref}
\usepackage{amsfonts}
\usepackage{amssymb}
\usepackage{dsfont}
\usepackage{mathrsfs}

\usepackage{capt-of,etoolbox}
\usepackage{graphicx}

\begin{document}


\title{Detective: An Attentive Recurrent Model for \\ Sparse Object Detection}



%

\author{Amine Kechaou\hspace{3em}Manuel Martinez\hspace{3em}Monica Haurilet\hspace{3em}Rainer Stiefelhagen\\
Institute for Anthropomatics and Robotics\\
Karlsruhe Institute of Technology, Karlsruhe, Germany\\
{\tt\small amine.kechaou@student.kit.edu} \\
{\tt\small \{manuel.martinez, haurilet, rainer.stiefelhagen\}@kit.edu} \\
}


\makeatother
\makeatletter
\let\@oldmaketitle\@maketitle
\renewcommand{\@maketitle}{\@oldmaketitle
  \centering
  \includegraphics[width=0.9\linewidth, height=4.2cm]{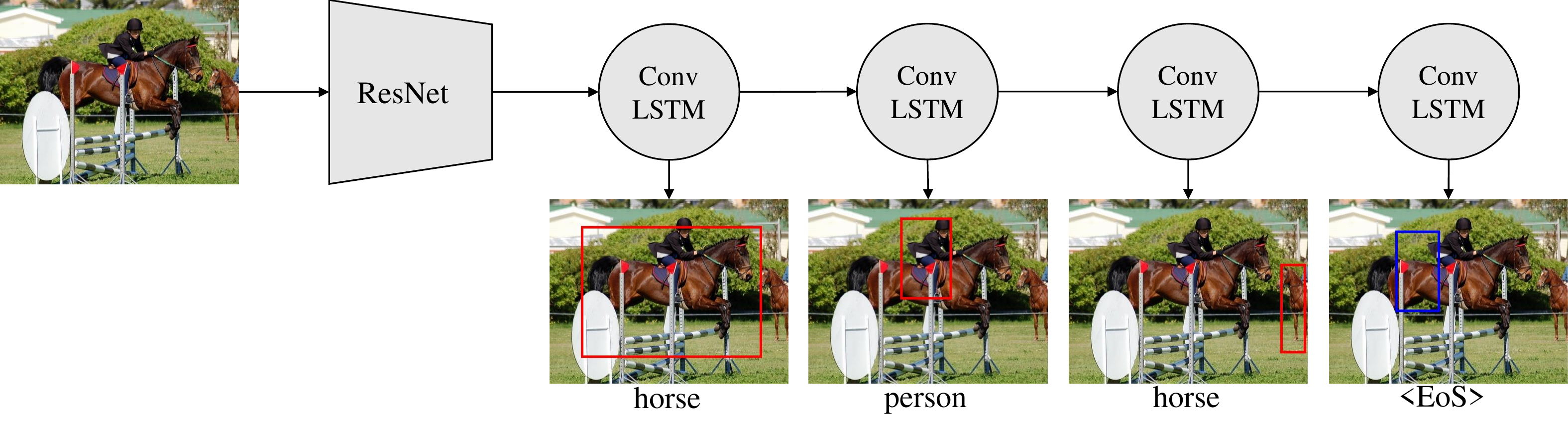}
\setcounter{figure}{0}
\vspace{-0.35cm}
\captionof{figure}{Overview of \textbf{Detective}: objects are detected one at a time until an End-of-Sequence (EoS) token is emitted.}
\label{fig:overview}
}

\makeatother

\maketitle

\begin{abstract}
In this work, we present Detective -- an attentive object detector that identifies objects in images in a \emph{sequential} manner.
Our network is based on an encoder-decoder architecture, where the encoder is a convolutional neural network, and the decoder is a convolutional recurrent neural network coupled with an attention mechanism. 
At each iteration, our decoder focuses on the relevant parts of the image using an attention mechanism, and then estimates the object's class and the bounding box coordinates. 
Current object detection models generate dense predictions and rely on post-processing to remove duplicate predictions.
Detective is a sparse object detector that generates a \emph{single} bounding box per object instance.
However, training a sparse object detector is challenging, as it requires the model to reason at the \emph{instance} level and not just at the class and spatial levels. 
We propose a training mechanism based on the Hungarian algorithm and a loss that balances the localization and classification tasks.
This allows Detective to achieve promising results on the PASCAL VOC object detection dataset.
Our experiments demonstrate that sparse object detection is possible and has a great potential for future developments in applications where the order of the objects to be predicted is of interest.
\end{abstract}





\section{Introduction}

In computer vision, the object detection task consists in localizing and categorizing objects of interest in a given image. 
It is one of the many tasks where deep learning based methods have achieved an outstanding performance \cite{ren2015faster, lin2017retinanet, law2018cornernet}. 
Current object detection architectures are trained to make \emph{dense} predictions, \ie, they generate a large amount of prospective detections, reaching even thousands of bounding boxes per image. 
Then, at inference time, the superfluous detections are filtered in a post-processing stage using Non-Maximum Supression (NMS).
The NMS algorithm greedily iterates over the generated detections, and, if two of them highly overlap with each other, it discards the one having a lower associated score. \looseness=-1

Although NMS is a very effective approach, it also comes with limitations. 
For example, if two objects in the image highly overlap with each other, and the network predicts one bounding box for each, NMS would usually wrongly discard one of them. 
The essence of this problem does not lie in NMS itself, but since object detectors generate dense predictions over each region in the image, they inherently do not reason about the objects at the instance level in the first place.
By densely predicting objects, conventional detectors also necessarily encounter a high class imbalance between object categories and the background class. 
This is due to the prevalence of background detections in comparison to foreground ones, since most regions in an image are part of the background. 
If not addressed, the class imbalance has a highly negative impact on the learning process.

To counter the aforementioned shortcomings, we introduce \emph{Detective} -- an attent\emph{ive} recurrent object \emph{detect}or that is fully end-to-end during training and inference, in the sense that it does not require any additional post-processing steps.
To this end, the model is explicitly designed for sparse object detection, and is trained to exactly predict one bounding box per object instance and jointly estimate the object's class.
It is based on an encoder-decoder architecture, where the encoder is a Convolutional Neural Network (CNN) and the decoder is a Convolutional Recurrent Neural Network (ConvRNN), specifically a Convolutional Long Short-Term Memory network (ConvLSTM). 
At inference time, the ConvLSTM predicts the objects in the image one at a time (\autoref{fig:overview}), until it decides that there are no more objects left to be predicted by emitting an End-of-Sequence (EoS) token.
After discarding the EoS token and the eventual background detections, the output of Detective represents the final set of detections, thus bypassing the need for any further post-processing steps.

Besides enabling a fully end-to-end learning for object detection by not requiring any post-processing, Detective comes with several other advantages.
In comparison to conventional object detectors, the training does not suffer from the class-imbalance problem that would otherwise occur when densely predicting bounding boxes over the whole image and as a result, extra steps like Online Hard Example Mining (OHEM) are not required.
The decoder in Detective is also lightweight by design since it leverages weight sharing across time in the RNN which makes it suitable for resource-restricted devices. 
Since objects are detected sequentially, the model can be leveraged to predict objects in a pre-defined order in one pass in cases where the order of the objects in the image matters.
Possible applications for ordered object detection can be found in fields like autonomous driving or assistance for the visually impaired. 
In autonomous driving, for example, it may be more critical to detect pedestrians at first before detecting vehicles and road signs. 
Thus, this key property of our framework has the potential to pave the way for \emph{ordered} object detection, where prioritizing different categories can lead to faster reactions of applied systems.

Our approach relies on RNNs, which have been successfully used in conjunction with CNNs in computer vision tasks like image captioning~\cite{vinyals2015show, xu2015show} where the output is expected to follow a certain order. 
The CNN-RNN framework has been used to a large extent in multi-label image classification \cite{ba2014, wang2016cnn_rnn, jin2016annotation}, but very few works have addressed the localization problem \cite{stewart2016people, romera2016ris, ren2017segatt} and even less works have explored the combination of the classification and localization tasks \cite{salvador2017rsis}. 
Since object detection is commonly framed as an unordered problem, the ground truth objects in the common object detection datasets are not subjected to any order. 
This mainly poses the challenge of how to assign target objects to the predictions made by the RNN during training. 
To address this, we build on related work \cite{stewart2016people} and conceive a training strategy that does not impose any given order on the RNN by dynamically matching the predictions to the target objects during training.
Finally, we evaluate our approach on the popular PASCAL VOC dataset~\cite{everingham2010pascal} that was used as a testbed for a variety of architectures for object detection. 
Therein, we show the capabilities of our approach for object detection, where we augment the evaluation with various analysis and settings of the Detective architecture.


\mypar{Contributions.}
A key contribution of this paper is the object detection framework that enables a fully end-to-end learning and inference without requiring any post-processing. 
Moreover, the Detective network opens the possibility for reasoning at instance level, and paves the way for \emph{ordered} object detection, enabling safety critical applications to prioritize object categories. 
We experimentally show that using an attention mechanism and a ConvLSTM positively impacts the localization capability of our model by augmenting spatial awareness. 


\section{Related Work}

\mypar{Two-Stage Object Detectors.} 
Driven by the advances of deep learning, the object detection task has experienced major progress in recent years. 
The first deep networks employed for object detection leverage classical region proposal algorithms (\eg, selective search~\cite{uijlings2013selective}) as a pre-processing step.
The found regions are then passed to a neural network that associate each of them with an object class or are marked as background and discarded.
A key network in this category was proposed by Girshick \etal called Region Convolutional Neural Network (R-CNN)~\cite{girshick2014rcnn}, which comprises a pre-processing step based on selective search to extract a fixed set of region proposals from the image. 
These regions are used to extract features from a pre-trained CNN, which are fed to Support Vector Machines (SVMs) for classification, while bounding box offsets are regressed using least mean squares to refine the localization given by the region proposals.
Despite outperforming prior works by a substantially large margin, R-CNN is slow due to the overhead caused by extracting a high number of object proposals that are fed repeatedly to the CNN. 

Several other extensions were introduced to counteract the drawbacks in the R-CNN model such as: (1) feeding the entire image to the CNN to extract the feature maps (SPP-Net~\cite{he2015spp}), 
(2)~extracting features using Region of Interest (RoI) pooling for the regions generated by selective search (Fast R-CNN~\cite{girshick2015fast}), (3)~replacing selective search with a Region Proposal Network (Faster R-CNN~\cite{ren2015faster}), (4) replacing the backbone network with VGG~\cite{ren2015faster}, FPN~\cite{lin2017FPN} or ResNet-FPN~\cite{he2017mask}.
Even though these network improve performance by a large margin compared to other approaches, two-stage object detectors have a relatively inferior speed.
This is mainly due to their complex architectures and the two-step process of extracting region proposals combined with their further processing.\looseness=-1 

\mypar{One-Stage Object Detectors.} 
Several networks were introduced that jointly regress the bounding box coordinates and estimate the object's class.
In YOLO \cite{redmon2016yolo}, the detection head jointly predicts bounding box offsets, class scores and objectness scores for each cell in the feature map estimated by the backbone.
To do so, each cell is responsible for detecting the class of an object, whose center falls into that grid cell.
In a similar vein, the Single Shot Detector (SSD)~\cite{liu2016ssd} estimates bounding boxes for each vector slice in the 3D tensor generated by the CNN.
However, instead of passing multiple scales of the image to the network, feature maps from different levels of the CNN backbone are concatenated and used to densely predict objects over the image. 
Multiple extensions were proposed based on these models: (1) inflation of the features for fine-grained detection~\cite{fu2017dssd}, (2) using DenseNet~\cite{iandola2014densenet} as a backbone without pre-training~\cite{shen2017dsod}, (3) balancing the class-wise distribution during training~\cite{shrivastava2016training,lin2017retinanet}.
Even though some of these networks are trained end-to-end, \emph{all} of them leverage a post-processing step to filter the strongly overlapping predicted boxes.

\mypar{Non-Maximum Suppression (NMS).} 
All previously introduced networks rely on Non-Maximum Suppression (NMS) as a post-processing step to filter potential duplicates, \ie, to obtain a single prediction per object instance. 
NMS successively traverses the predictions in a greedy fashion and if two bounding boxes strongly overlap (\ie, have a high intersection over union), the box with the lower assigned confidence score is discarded. 
The overlap threshold to merge the bounding boxes is handled as a conventional hyper-parameter that is set depending on the dataset and model.
While the final outputs of previously proposed frameworks for object detection are mostly free of duplicates, the raw outputs are often not, as the models are not explicitly trained to output only \emph{one} bounding box per object instance. 
Hosang \textit{et al.} propose a learnable module that aims to replace NMS \cite{hosang2017nms}.
The network learns new scores for the detections made by the object detector, so that for each object only one detection would have a high score, and all others would have a very low score, thus, can be discarded by thresholding. 
Nonetheless, whether with greedy or learnable NMS, the underlying object detector still generates a high number of detections that need to be filtered afterwards. 
In comparison to these approaches, we aim to remove the filtering step altogether and only want to directly generate a \emph{single} bounding box per object instance, thus allowing for a reasoning on the instance level. 

\mypar{Recurrent Neural Networks (RNNs).} 
In computer vision, RNNs have been extensively used in image captioning~\cite{karpathy2015deep, vinyals2015show, xu2015show, rennie2017self}, where a CNN first computes an image representation and then the RNN (often coupled with an attention mechanism~\cite{xu2015show, rennie2017self}) generates a description of the image one word at a time.
RNNs have also been used to some extent in multi-label image classification~\cite{ba2014, wang2016cnn_rnn, jin2016annotation}, where an image is associated with a variable number of labels and the RNN predicts these one at a time. 
To a lesser degree, RNNs were employed in the localization task to localize object instances in an image sequentially. 
Most relevant is the work by Stewart \textit{et al.} for people detection~\cite{stewart2016people}, where a CNN first encodes an image into feature maps, and a grid of LSTM networks sequentially predict bounding boxes at each location in the feature maps until a stopping condition is met. 
During training, the ground truth objects are matched to the predictions using the Hungarian algorithm~\cite{kuhn1955hungarian}. 
At inference time, the model relies on a stitching process also based on the Hungarian algorithm to filter the duplicate predictions. 
Our work is similar to the work by Stewart \textit{et al.} in the way that it uses the Hungarian algorithm in the matching process and extends it to object detection in general. 
However, it differs substantially from an architectural standpoint as it uses a single ConvLSTM instead of a grid of LSTM networks and does not require any post-processing at inference time. 
In a similar vein to our work, some other works also build on the work by Stewart \textit{et al.} and extend it for instance segmentation~\cite{romera2016ris, ren2017segatt, salvador2017rsis}.
\looseness=-1


\begin{figure}[t!]
\centering
\includegraphics[width=.5\textwidth,keepaspectratio]{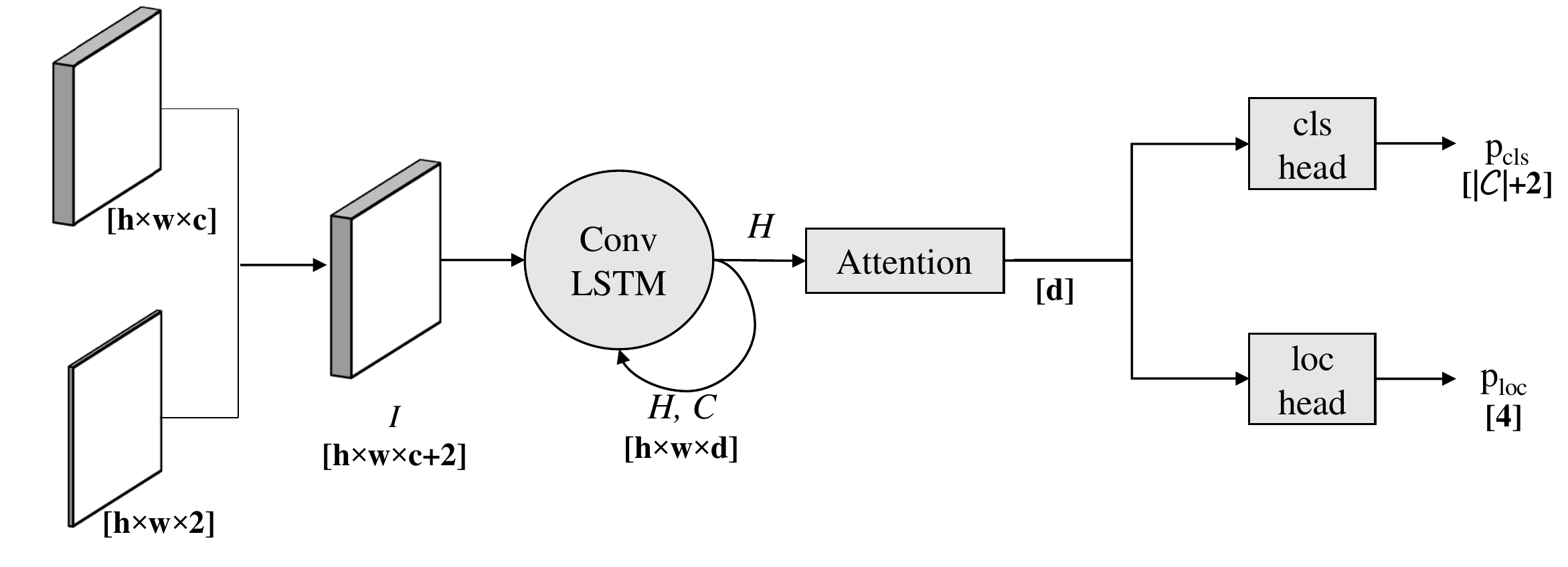}
\vspace{-0.725cm}
\caption{Overview of Detective's architecture. The feature maps generated by the ConvLSTM are dynamically pooled by an attention module. 
}
\label{fig:convatt}
\end{figure}

\section{Model Architecture}
Detective is based an encoder-decoder architecture, where the encoder is a CNN and the decoder consists of a Convolutional LSTM (ConvLSTM)~\cite{xingjian2015convlstm} coupled with an attention mechanism.
Given an input image, the encoder computes an image representation in form of a 3D tensor that is then passed to the decoder.
At each iteration, the ConvLSTM-based decoder estimates a pair of class probabilities and location offsets. 
While the number of iterations is fixed during training (Section~\ref{sec:training}), at inference time, the ConvLSTM stops iterating when the End-of-Sequence (EoS) token is generated. 
Thus, the model has to learn both to \emph{detect objects} and to \emph{stop iterating} when there are no more objects left in the scene.



\subsection{Encoder}
For the encoder network, we leverage a ResNet50 model~\cite{he2016resnet} pre-trained on ImageNet \cite{deng2009imagenet} as it has shown to be the backbone of choice in many modern object detectors~\cite{dai2016rfcn, lin2017FPN, lin2017retinanet}.
Therein, we remove the average-pool and fully-connected layers, and obtain a final visual representation of the size $h\times w\times c$.
Although the extracted feature maps can be directly passed to the decoder, we ease the learning of the bounding boxes by explicitly encoding positional information and fusing them with the semantic representations. 
More formally, we concatenate the extracted feature maps with positional embedding matrices  $F^x, F^y \in \mathbb{R}^{h\times w}$ in both the $x$- and $y$-dimensions as follows:
\begin{equation}
F_{{i, j}}^x = j, F_{{i, j}}^y = i \text{ where }  \newline i = 0,\dots, h - 1, j = 0,\dots, w - 1.
\end{equation}
After the merging operation, we obtain a 3D tensor $I$ of the size $h\times w\times (c + 2)$, which is passed to the decoder network.

\subsection{Decoder}
\mypar{ConvLSTM. }
To generate the sequence of bounding boxes and associated object classes, we employ a recurrent neural network, specifically a ConvLSTM~\cite{xingjian2015convlstm}. 
In contrast to conventional LSTM networks, the input, the hidden state and the cell state are 3D tensors instead of vectors. Consequently, the matrix-multiplications with the weight matrices is replaced by convolutions. 
The ConvLSTM is implemented following Equation~\ref{eq:convlstm}, where at iteration $ t$, $ X_t $ is the input (set to the feature maps $ I $), $H_t$ is the hidden state, $C_t $ is the cell state and $ i_t, f_t $ and $ o_t$ are respectively the input, forget and output gates. 
The convolution operation is represented by $ \ast $, while $ \circ $ denotes the element-wise product. 

\begin{equation}
\label{eq:convlstm}
\begin{aligned}
i_t &= \sigma (W_{xi} \ast X_t + W_{hi} \ast H_{t - 1} + W_{ci} \circ C_{t - 1} + b_{i}) \\
f_t &= \sigma (W_{xf} \ast X_t + W_{hf} \ast H_{t - 1} + W_{cf} \circ C_{t - 1} + b_f ) \\
C_t &= f_t  \circ C_{t - 1} + i_t \circ tanh(W_{xc} \ast X_t + W_{hc} \ast H_{t - 1} + b_c) \\
o_t &= \sigma (W_{xo} \ast X_t + W_{ho} \ast H_{t - 1} + W_{co} \circ C_t + b_o) \\
H_t &= o_t \circ tanh(C_t)
\end{aligned}
\end{equation}

\mypar{Attention-based dynamic pooling. }
The hidden state at each iteration step acts as a representation of the object currently being detected. 
To decode this representation into a pair of class probabilities and location offsets, an attention mechanism dynamically pools the hidden state at each iteration step into a vector, which is subsequently passed through the classification and localization branches. 
The attention mechanism consists of a neural network with a single hidden layer with $ d $ neurons and an output layer with a single neuron. 
Given a hidden state  $ H_i \in \mathbb{R}^{h\times w\times d} $, the neural network produces an attention map $ A \in \mathbb{R}^{h\times w\times 1} $. 
The values of the soft attention map are obtained by applying the softmax normalization on $ A$ modeling a probability distribution over each location being relevant to generate the current object.
The final output $ o_i \in \mathbb{R}^d $  is the result of the dot product between the hidden state and the attention weights given by:
\begin{equation}
\label{eq_output}
o_i = H_i^f \cdot \softmax(A^f_i),
\end{equation}
where $ H_i^f \in \mathbb{R}^{d\times (h \cdot w)} $ and $ A^f_i \in \mathbb{R}^{h \cdot w} $ are flattened versions of $ H_i $ and $ A_i $ along the spatial dimensions.
The softmax function $ \softmax : \mathbb{R}^{k} \to \mathbb{R}^{k}$  normalizes the input over all its values as follows:

\begin{equation}
\softmax(z)_i = \frac{e^{z_i}}{\sum_{j=1}^{k} e^{z_j}} \text{ for } i = 1,..., k \text{ and } z \in \mathbb{R}^{k},
\end{equation}
where $k\in \mathbb{N}$ is a fixed and finite number, which we set in Equation~\ref{eq_output} to: $k = h \cdot w$. 

\begin{figure}[t!]
\centering
\includegraphics[width=.5\textwidth,keepaspectratio]{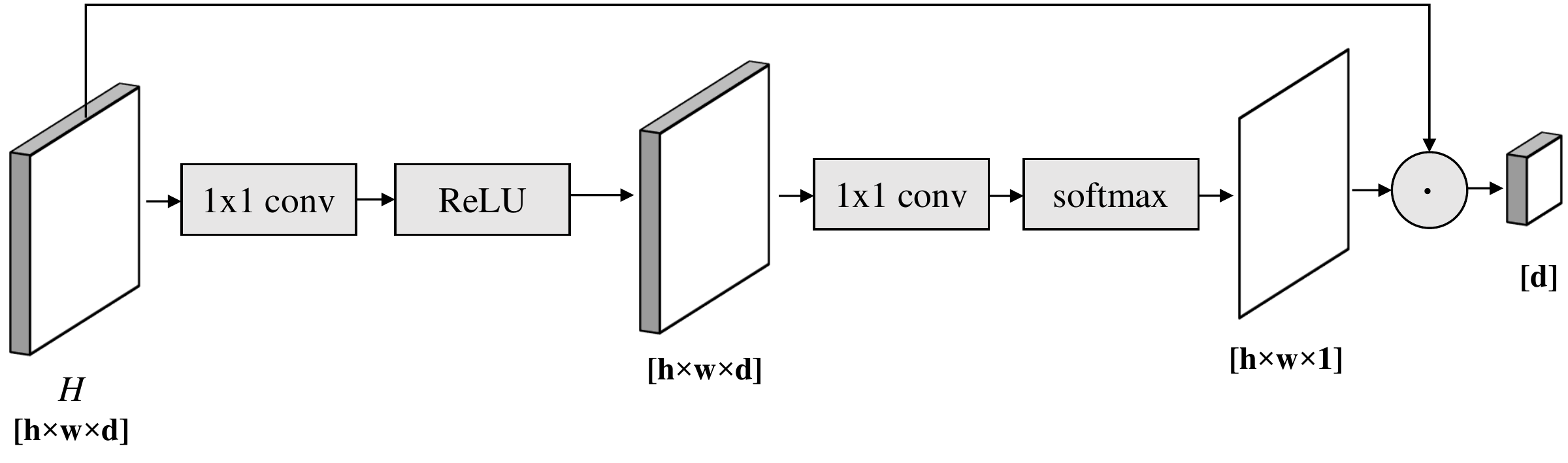}
\caption{The attention mechanism in Detective. 
The attention map is obtained from two convolution layers followed by softmax normalization. 
The final vector representation of the image is generated by weighting the input 3D tensor with the values in the attention map.}
\label{fig:attdec}
\vspace{-0.45cm}
\end{figure}

\mypar{Classification and localization branches. }
The classification and localization branches consist of simple linear layers followed by the softmax and sigmoid normalization functions, respectively, to compute the class probability scores and the bounding box offsets. 
Since the model needs to signal when to stop iterating, the set of object classes $ \mathcal{C} $ is extended with an extra class representing the End-of-Sequence (EoS) token. 
In addition, an extra background class is used to classify predictions that do not sufficiently overlap with any ground truth objects. 
At inference time, such predictions are simply discarded. 
Concretely, given an object representation $  o \in \mathbb{R}^d  $, the classification scores are computed as:
\begin{equation}\label{eq:cls_head}
p_{cls} = \softmax(W_{cls}^\intercal \cdot o + b_{cls}),
\end{equation}
while the bounding box offsets are computed as follows: 
\begin{equation}\label{eq:loc_head}
p_{loc} = \sigma(W_{loc}^\intercal \cdot o +  b_{loc})
\end{equation}
where $ W_{cls} \in \mathbb{R}^{d\times (|\mathcal{C}| + 2)} $ and $ W_{loc} \in \mathbb{R}^{d\times 4}, b_{cls} \in \mathbb{R}^{|\mathcal{C}| + 2} $ and $ b_{loc} \in  \mathbb{R}^4$ are learnable parameters and $ \sigma $ is the sigmoid function applied element-wise, which is defined as:
$$ \sigma: \mathbb{R} \to [0, 1], \sigma(x) = \frac{1}{1 + e^{-x}}. $$
We apply the sigmoid function since the bounding box offsets are computed relative to the top-left corner in the image, and thus, it needs to be bound in the $ [0, 1] $ range.

\section{Learning Procedure}
\label{sec:training}
During training, the model needs to learn to \emph{localize} and \emph{recognize} the objects in the scene as well as to learn when to \emph{stop iterating}, \ie when there are no more objects of interest left in the image. 
Our RNN is trained in a supervised fashion on conventional object detection datasets comprising images with associated labels of the object class and bounding box coordinates. 
In contrast to common object detectors, a key challenge in our setup is that there is no straightforward way to map the predicted instances to the ground truth object annotations. 
While RNNs necessitate \emph{sequential} data, the object detection task comprises an unordered set of object instances present in the image.
Thus, a major aspect for our framework design is how to map the \emph{set of object bounding boxes} to the \emph{predictions of our RNN}.   
To this end, we propose to leverage the popular Hungarian Algorithm~\cite{kuhn1955hungarian} for establishing the matching between our predictions and the ground truth.

\subsection{Dynamic Matching}
Since object detection is usually not treated as a sequential task, the ground truth objects in an image are unordered and, thus, a direct mapping between the ground truth objects set and the predictions does not exist. 
Several works using RNNs for object localization \cite{stewart2016people} or instance segmentation  \cite{romera2016ris, ren2017segatt} address this problem by framing it as an assignment problem in a bipartite graph, where one part is the set of ground truth objects and the other is the set of predictions. 
The weight on each edge on the graph represents the cost of matching a ground truth object to a prediction.

In our problem setup, the space of target labels can be expressed as $ \mathcal{S}_t \subset \mathds{1}^{|\mathcal{C}| + 2} \times [0, 1]^4 $. 
We denote the set of $ n $ target labels in one image by $ \mathcal{T} \subset {\mathcal{S}_t}^n $. 
A single target label is defined as the tuple $ (t_{cls}, t_{loc}) \in \mathcal{T} $, where $ t_{cls} $ is the one-hot encoded target class and $ t_{loc} $ is the vector of target bounding box offsets. 
We set the number of RNN iterations during training to $ m = n + 1 $, where $ n $ is the number of ground truth objects. 
While the first $ n $ predictions are matched to ground truth objects, the last prediction is matched to the EoS token.
The prediction space softened by the softmax normalization is expressed as $ \mathcal{S}_p \subset [0, 1]^{|\mathcal{C}| + 2} \times [0, 1]^4 $. 
We denote the set of $ n $ predictions for one image with $ \mathcal{P} \subset {\mathcal{S}_p}^{n} $. 
Note that only the first $ m - 1 = n $ predictions are matched to the ground truth objects as the $ m $-th prediction is assigned to the EoS token. 
Similarly as for the ground truth labels, a single prediction of our model is a tuple $ (p_{cls}, p_{loc}) \in \mathcal{P} $, where $ p_{cls} $ is the vector of class probability scores and $ p_{loc} $ is the vector of predicted bounding box offsets.

Let  $ f : \mathcal{T} \times \mathcal{P} \to \mathbb{R}^{n \times {n}} $ be the cost function between the target and the prediction, which is defined as the weighted sum of the classification and localization losses:
\begin{multline} \label{matching_cost_function}
f(t, p) = \mu_{cls}\mathcal{L}_{cls}(t_{cls}, p_{cls}) + \mu_{loc}\mathcal{L}_{loc}(t_{loc}, p_{loc}),
\end{multline} 
where $ \mathcal{L}_{cls} $ and $ \mathcal{L}_{loc} $ are the classification and localization loss function, respectively, and $ \mu_{cls} $ and $ \mu_{loc} $ are hyper-parameters. 
The goal is then to find a bijective matching function 
$g: \mathcal{T} \to \mathcal{P}$
such that the overall matching cost $ \sum_{t \in \mathcal{T}} f(t, g(t)) $ is minimal. 
This frames the assignment problem as a minimization problem, whose optimal solution can be found in polynomial time with a complexity in $ \mathcal{O}(n^3) $ by means of the Hungarian algorithm \cite{kuhn1955hungarian}.

\subsection{Loss Function}
Given the bijective mapping $ g $ between ground truth objects and predictions, let $ \mathcal{M} = \{(t, p) \in \mathcal{T} \times \mathcal{P}\ |\ p = g(t)\ \forall t \in \mathcal{T}\} $ be the set of the matched pairs of targets and predictions. 
We define the loss to be minimized as a weighted sum of the classification and localization losses:
\begin{equation} \label{eq:loss}
L = \lambda_{cls}L_{cls} + \lambda_{loc} L_{loc}.
\end{equation}

The classification loss is responsible for learning to classify objects, discard boxes as  background and to signal when to stop iterating. 
Following works like Faster R-CNN~\cite{ren2015faster}, predictions whose Intersection over Union (IoU) with their matched target objects is greater than $0.5$ are considered as foreground predictions, while the ones with an IoU less than $0.3$ are marked as background. 
Predictions with an IoU between $0.3$ and $0.5$ with their matched target objects are ignored when computing the classification loss. 
The set of pairs of targets and foreground predictions is denoted by:
\begin{equation}
\mathcal{M}_{FG} = \{(t, p) \in M\ |\ IoU(t, p) \geq 0.5\}
\end{equation}
and the set of pairs of targets and background predictions is depicted by:
\begin{equation}
\mathcal{M}_{BG} = \{(t, p) \in M\ |\ IoU(t, p) < 0.3\}.
\end{equation}
This allows us to define $ L_{cls} $ as 
\begin{multline}\label{eq:loss_cls}
L_{cls} =  \sum_{(t,p)\in \mathcal{M}_{FG}}{\mathcal{L}_{cls}(t_{cls}, p_{cls})} \\ 
+ \sum_{(t,p)\in \mathcal{M}_{BG}}{\mathcal{L}_{cls}(BG, p_{cls})} + \mathcal{L}_{cls}(EoS, p^{(m)}_{cls}),
\end{multline}
\noindent
where $ \mathcal{L}_{cls} $ is the negative log-likelihood loss function, $ BG $ is the background class and $ p^{(m)} $ is the last prediction produced by the RNN.

The localization loss $ L_{loc} $ is computed over all target prediction pairs in $\mathcal{M}$:
\begin{equation} \label{eq:loss_loc}
L_{loc} = \sum_{(t,p)\in \mathcal{M}}{\mathcal{L}_{loc}(t_{loc}, p_{loc})},
\end{equation} 
\noindent
where $ \mathcal{L}_{loc} $ is the L2 loss function. 
Note that the offset regression is performed only on the first $ n $ predictions, regardless of the IoU between these and their associated targets.
Since the number of steps the RNN performs during inference is not given beforehand, the network needs to learn when the total number of objects is reached, \ie, has to stop iterating.

\subsection{Parameter setup}
We learn the parameters by minimizing our loss function (Equation~\ref{eq:loss}) using Adam~\cite{kingma2014adam} with an initial learning rate of $ 10^{-4} $, an exponential decay for the first moment of $ 0.9 $ and for the second moment of $ 0.999 $. 
The matching and loss weights are experimentally set to $ \mu_{loc} = \lambda_{loc} = 16$, $\mu_{cls} = 0$ and $\lambda_{cls} = 0.1$.
The ConvLSTM in Detective comprises  $ 512 $ filters of size $ 3 \times 3 $, stride $ S = 1 $ and \textit{same} padding to preserve the spatial dimensions.



\begin{figure*}[t!]
\centering

\begin{minipage}{.49\textwidth}
\centering
\begin{tikzpicture}
\tikzstyle{every node}=[font=\footnotesize]
\begin{axis}[
    title={Predictions' distribution for Detective (IoU$\geq$0.5)},
    height=4.5cm,  width=7.00cm,
    smooth,
    stack plots=y,
    area style,
    xtick = {1,2,3,4,5,6,7,8,9},
    xmin=1,xmax=9,ymin=0,ymax=1,
    xlabel=ConvLSTM iteration,
    legend pos={outer north east},
    enlargelimits=false]
\addplot+[green,fill=green!30!white] plot coordinates { (1, 0.753231) (2, 0.291195) (3, 0.250856) (4, 0.181756) (5, 0.279182) (6, 0.116667) (7, 0.148387) (8, 0.0904255) (9, 0.165289) } \closedcycle;
\addplot+[red,fill=red!30!white] plot coordinates { (1, 0.218901) (2, 0.188005) (3, 0.234869) (4, 0.321674) (5, 0.274368) (6, 0.377778) (7, 0.383871) (8, 0.425532) (9, 0.504132) } \closedcycle;
\addplot+[orange,fill=orange!30!white] plot coordinates { (1, 0) (2, 0.0234249) (3, 0.0380662) (4, 0.0336077) (5, 0.0493381) (6, 0.0351852) (7, 0.0387097) (8, 0.0265957) (9, 0.0413223) } \closedcycle;
\addplot+[gray,fill=gray!30!white] plot coordinates { (1, 0.0278675) (2, 0.0278675) (3, 0.0312143) (4, 0.0329218) (5, 0.0469314) (6, 0.0444444) (7, 0.0354839) (8, 0.101064) (9, 0.0495868) } \closedcycle;
\addplot+[red,fill=red!30!white,postaction={pattern=north west lines, pattern color=red}] plot coordinates { (1, 0) (2, 0.137722) (3, 0.199848) (4, 0.294239) (5, 0.292419) (6, 0.368519) (7, 0.380645) (8, 0.351064) (9, 0.22314) } \closedcycle;
\addplot+[green,fill=green!30!white,postaction={pattern=north west lines, pattern color=green}] plot coordinates { (1, 0) (2, 0.331785) (3, 0.245147) (4, 0.135802) (5, 0.0577617) (6, 0.0574074) (7, 0.0129032) (8, 0.00531915) (9, 0.0165289) } \closedcycle;
\legend{{good},{bad},{duplicate},{background},{EoS:incorrect},{EoS:correct}}
\end{axis}
\end{tikzpicture}
\end{minipage}
~
\begin{minipage}{.49\textwidth}
\centering
\begin{tikzpicture}
\tikzstyle{every node}=[font=\footnotesize]
\begin{axis}[
    title={Predictions' distribution for Detective (IoU$\geq$0.25)},
    height=4.5cm,  width=7.00cm,
    smooth,
    stack plots=y,
    area style,
    xtick = {1,2,3,4,5,6,7,8,9},
    xmin=1,xmax=9,ymin=0,ymax=1,
    xlabel=ConvLSTM iteration,
    legend pos={outer north east},
    enlargelimits=false]
\addplot+[green,fill=green!30!white] coordinates { (1, 0.852585) (2, 0.358845) (3, 0.33308) (4, 0.288066) (5, 0.336943) (6, 0.227778) (7, 0.245161) (8, 0.265957) (9, 0.347107) } \closedcycle;
\addplot+[red,fill=red!30!white] coordinates { (1, 0.119548) (2, 0.0932956) (3, 0.108108) (4, 0.127572) (5, 0.132371) (6, 0.159259) (7, 0.154839) (8, 0.12766) (9, 0.214876) } \closedcycle;
\addplot+[orange,fill=orange!30!white] coordinates { (1, 0) (2, 0.0504847) (3, 0.0826037) (4, 0.121399) (5, 0.133574) (6, 0.142593) (7, 0.170968) (8, 0.148936) (9, 0.14876) } \closedcycle;
\addplot+[gray,fill=gray!30!white] coordinates { (1, 0.0278675) (2, 0.0278675) (3, 0.0312143) (4, 0.0329218) (5, 0.0469314) (6, 0.0444444) (7, 0.0354839) (8, 0.101064) (9, 0.0495868) } \closedcycle;
\addplot+[red,fill=red!30!white,postaction={pattern=north west lines, pattern color=red}] coordinates { (1, 0) (2, 0.120557) (3, 0.162543) (4, 0.243484) (5, 0.251504) (6, 0.307407) (7, 0.335484) (8, 0.31383) (9, 0.214876) } \closedcycle;
\addplot+[green,fill=green!30!white,postaction={pattern=north west lines, pattern color=green}] coordinates { (1, 0) (2, 0.34895) (3, 0.282451) (4, 0.186557) (5, 0.0986763) (6, 0.118519) (7, 0.0580645) (8, 0.0425532) (9, 0.0247934) } \closedcycle;
\legend{{good},{bad},{duplicate},{background},{EoS:incorrect},{EoS:correct}}
\end{axis}
\end{tikzpicture}
\end{minipage}

\caption{
Distribution of the predictions generated by Detective at each ConvLSTM iteration.
A \textit{good} prediction has an IoU$\geq$0.5 (left) or IoU$\geq$0.25 and is correctly classified.
A \textit{duplicate} prediction identifies an object that was already predicted. 
A \textit{bad} prediction does not match any ground truth object (wrong class or IoU lower than the threshold).
\textit{Background} predictions are labeled as background by the network and are discarded at inference time.
An EoS token is \textit{correct} if all ground truth objects have already been detected, and otherwise \textit{incorrect}.
}
\label{fig:token_distribution}

\end{figure*}
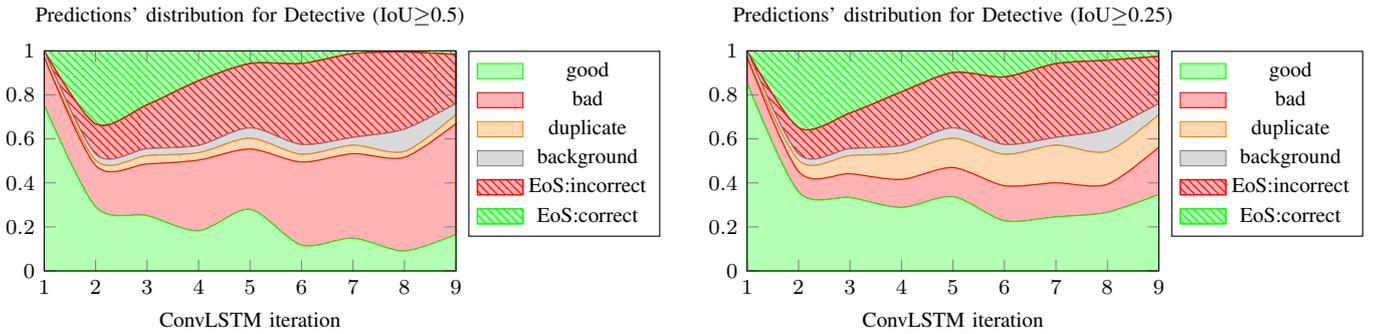

\section{Evaluation}
The goal of this work is to achieve end-to-end object detection that does not require any post-processing at inference time.  
As it is a common practice in the field of object detection, we perform the experiments on the PASCAL VOC \cite{everingham2010pascal} dataset, which contains thousands of images that were annotated with bounding boxes along with their corresponding classes. 
PASCAL VOC comes in two versions: VOC07 and VOC12, both of which are split into three subsets: \texttt{train}, \texttt{val} and \texttt{test}, and have annotations of $20$ classes. 
For VOC07, the annotations of the \texttt{test} set are publicly available, whereas for VOC12, only the unannotated images from the \texttt{test} set are available and the annotations are kept private on an evaluation server. 
We follow other works \cite{ren2015faster, dai2016rfcn, liu2016ssd} and use the combination of the \texttt{train} and \texttt{val} sets from VOC07 and VOC12 (\texttt{trainval07+12}) to train our model and validate on the \texttt{test} set of VOC07 (\texttt{test07}). 
Detective achieves an mAP of $52.0$\% on \texttt{test07} without requiring any post-processing. 
As a reference, other object detectors that rely on NMS in a post-processing stage like R-FCN \cite{dai2016rfcn} or DSSD \cite{fu2017dssd} achieve $80.5$\% and $81.5$\% respectively on \texttt{test07}.  

\subsection{Ablation Study}
To assess the contribution of the different elements constituting our model, we conduct an ablation study where the encoder is kept the same throughout all experiments and different components of the decoder are omitted. 
The different design decisions that we evaluate are the choice of a convolutional LSTM network instead of a 1D LSTM network, the attention mechanism, the additional positional embeddings and the additional background class.
We choose as a baseline a model where all these components are turned off, that is, a simple LSTM network whose input is an average-pooled image feature vector without additional positional encodings and without background classification. 
The results of this experiment are reported in \autoref{table:ablation}, while the per-class average precision resulting from the different models is reported in \autoref{table:ap}. \looseness=-1

\begin{table}[t!]
\caption{Results of Detective under different decoder settings} 
\label{table:ablation}
\scriptsize
\centering
\begin{tabular}{@{} c | c  c  c  c | c @{}}
\toprule
Model & \thead{\scriptsize Conv- \\ \scriptsize LSTM} & \thead{\scriptsize Attention} & \thead{\scriptsize Positional \\ \scriptsize embeddings} & \thead{\scriptsize Background \\ \scriptsize classification} & \thead{ \scriptsize mAP (\%) \\ \scriptsize on VOC07} \\
\midrule
Baseline & \red{\xmark} & \red{\xmark}  & \red{\xmark}  & \red{\xmark}  & 40.7 \\
\midrule
(A) & \red{\xmark} & \green{\cmark} & \green{\cmark} & \green{\cmark} & 49.4 \\

(B) & \green{\cmark} & \red{\xmark}  & \green{\cmark} & \green{\cmark} & 46.5 \\

(C) & \green{\cmark} & \green{\cmark} & \red{\xmark}  & \green{\cmark} & 47.9 \\

(D) & \green{\cmark} & \green{\cmark} & \green{\cmark} & \red{\xmark}  & 49.9 \\
\midrule
Detective & \green{\cmark} & \green{\cmark} & \green{\cmark} & \green{\cmark} & \textbf{52.0} \\
\bottomrule
\end{tabular}
\end{table}

\begin{table*}[t]
\tiny
\centering
\caption{mAP and per-class average precision (\%) on \texttt{test07}}
\label{table:ap}
\begin{tabular}{c | c | c  c  c  c  c  c  c  c  c  c  c  c  c  c  c  c  c  c  c  c} 
\toprule
~Model~ & ~mAP~ & ~plane~  & ~bike~ & ~bird~ & ~boat~  & ~bottle~ & ~bus~ &  ~car~  & ~cat~ & ~chair~ & ~cow~ & table & dog & horse & mbike & person & plant & sheep & sofa & train & tv \\
\midrule
~Baseline~ & 40.7 & 54.3 & 54.3 & 39.0 & 31.2 & 5.8 & 51.7 & 43.3 & 68.0 & 11.1 & 34.1 & 43.1 & 59.9 & 69.6 & 52.5 & 33.6 & 7.9 & 24.0 & 44.8 & 68.4 & 17.2 \\ 
(A) & 49.2 & 55.0 & 59.6 & 48.4 & 34.2 & 14.4 & 60.0 & 56.8 & \textbf{76.5} & 20.7 &40.4 & 50.4 & 68.6 & 74.8 & 64.1 & 46.3 & 15.5 & 41.3 & 50.4 & 72.4 & 35.0 \\

(B) & 46.5 & 55.6 & 56.7 & 43.3 & 35.9 & 6.2 & 59.5 & 50.8 & 75.7 & 18.8 & 43.2 & 47.0 & 63.2 & 73.6 & 57.9 & 38.4 & 10.9 & \textbf{45.2} & 43.3 & 73.9 & 31.2 \\

(C) & 47.8 & 54.9 & 57.8 & 43.9 & 37.8 & \textbf{15.5} & 57.1 & 55.1 & 69.7 & 18.5 & 41.9 & 55.6 & 63.2 & 75.6 & 61.9 & 42.9 & 15.0 & 35.5 & 49.9 & 73.6 & 33.0  \\

(D) & 49.9 & 58.0 & 60.2 & 51.2 & \textbf{42.1} & 13.1 & \textbf{61.5} & 56.1 & 75.6 & 18.7 & 44.5 & 52.1 & 65.9 & 73.9 & 64.0 & 43.0 & 16.3 & 40.0 & 51.2 & \textbf{75.7} & 36.5 \\
Detective & \textbf{52.0} & \textbf{60.8} & \textbf{64.2} & \textbf{52.7} & 39.6 & 14.1 & \textbf{59.3} & \textbf{59.2}  & 75.3 & \textbf{22.7}  & \textbf{45.8} & \textbf{53.1} & \textbf{71.8} & \textbf{76.8} & \textbf{68.1} & \textbf{47.3} & \textbf{18.4} & 40.7 & \textbf{54.0} & 74.6 & \textbf{42.0} \\
\bottomrule
\end{tabular}
\vspace{-0.45cm}
\end{table*}

The different design decisions bring an overall improvement of around $11$\% in the mAP over the baseline. 
In \autoref{table:ablation}, we mark with model (A) a 1D LSTM network whose input is an attended feature vector conditioned on both the feature maps and the hidden state.
Model (A) improves the mAP score by more than $2$\% in comparison to the baseline approach that does not make use of convolutional layers. 
In model (B), we switch the attention-based dynamic pooling a simple average pooling.
The attention mechanism has the most significant impact in Detective by bringing an improvement of $5.5$\% in mAP, which confirms that attention plays indeed a critical role in inducing spatial awareness.
To assess the role of the positional embeddings, we omit the concatenation of these with the feature maps resulting from the encoder in model (C) and compare its performance with Detective. 
We find that the additional positional embeddings play an important role in helping the model in the localization task as they improve the performance of our model by nearly $4$\%. 
When trained without classifying background detections as such, model (D) in \autoref{table:ablation} achieves a lower mAP score than Detective by about $2$\%. 
Even though background classification in Detective is not as crucial as it is the case in other object detectors, for bounding boxes are in our case not predicted at each location in the image, it still improves performance by more than $2$\%.


\subsection{Performance Analysis}

\begin{figure}[t]
\centering
\begin{tikzpicture}
\tikzstyle{every node}=[font=\footnotesize]
\begin{axis}[
    height=4.5cm,  width=7.00cm,
    smooth,
    mark=none,
    xtick = {1,2,3,4,5,6,7,8,9},
    xmin=1,xmax=9,ymin=0,ymax=100,
    xlabel=ConvLSTM iteration,
    ylabel=Precision (\%),
    legend pos={outer north east},
    enlargelimits=false]

\addplot+[mark=none] coordinates { (1, 77.4823) (2, 57.9349) (3, 47.8924) (4, 33.8442) (5, 46.3074) (6, 22.028) (7, 25.9887) (8, 16.6667) (9, 23.2558) };

\addplot+[mark=none] coordinates { (1, 70.6583) (2, 32.4691) (3, 36.4248) (4, 19.9516) (5, 27.0541) (6, 10.9541) (7, 16.2162) (8, 12.5984) (9, 18.5185) };

\legend{Detective,baseline}
\end{axis}
\end{tikzpicture}

\caption{Precision of the detections at each iteration}
\label{fig:precision_iteration}
\end{figure}
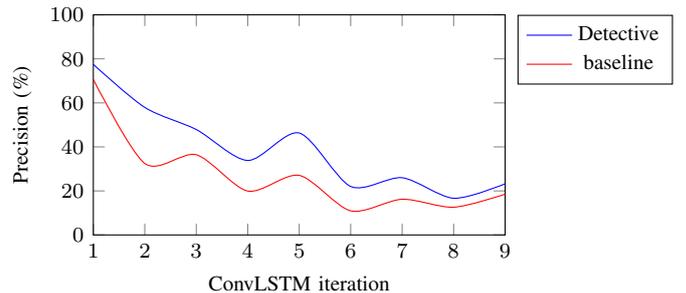


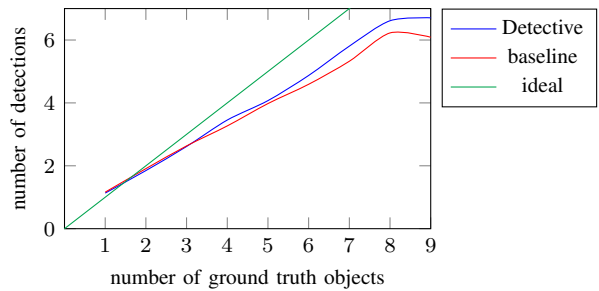
\begin{figure}[t]
\centering

\begin{tikzpicture}
\tikzstyle{every node}=[font=\footnotesize]
\begin{axis}[
    height=4.5cm,  width=6.40cm,
    smooth,
    mark=none,
    xtick = {1,2,3,4,5,6,7,8,9},
    xmin=0,xmax=9,ymin=0,ymax=7,
    ylabel=number of detections,
    xlabel=number of ground truth objects,
    legend pos={outer north east},
    enlargelimits=false]

\addplot+[mark=none] coordinates { (1, 1.13181) (2,1.84768 ) (3, 2.61439) (4, 3.45531) (5, 4.07292) (6, 4.87261) (7,5.80519 ) (8, 6.6129) (9, 6.70588) };

\addplot+[mark=none] coordinates { (1, 1.16175 ) (2,1.90977 ) (3, 2.63284) (4, 3.26816) (5,3.97917 ) (6,4.59236  ) (7,5.32468 ) (8,6.22581) (9, 6.08824 ) };

\addplot[Green] coordinates { (-1, -1 ) (19, 19) };

\legend{Detective,baseline, ideal}
\end{axis}
\end{tikzpicture}

\caption{Number of detections vs. number of ground truth objects. 
The green line shows the ideal case where the number of detected objects matches the number of ground truth objects. 
Our networks tend to make less detections than there are ground truth objects. Nonetheless, Detective shows better performance than the baseline.
}
\label{fig:detections_vs_labels}
\end{figure}

We investigate the distribution of the predictions generated by Detective over several ConvLSTM iterations in \autoref{fig:token_distribution}.
We observe that the first prediction is generated with a high degree of precision, after which the performance degrades. By lowering the IoU threshold at which detections are accepted from the standard to $0.5$ to $0.25$, we find that the number of true positives significantly increases, which suggests that Detective's performance is especially limited by its localization ability.
Nevertheless, we can observe in \autoref{fig:precision_iteration} that Detective's performance at each iteration is steadily better than the baseline's, which is mainly attributed to the attention mechanism and the use of a ConvLSTM. This is also demonstrated in \autoref{fig:detections_vs_labels}, which shows that the number of detected objects is highly correlated with number of ground truth objects. 
While Detective generates less detections when the number of ground truth objects is high, it consistently improves over the baseline.

The small number of predictions labeled as background as seen \autoref{fig:token_distribution} reflects the reason why classifying background predictions comes with an improvement of only 2\% (\autoref{table:ablation}). 
Importantly, this attests to how our learning procedure yields a significantly lower number of background detections in comparison to common object detectors. 
Hence, the training inherently does not suffer from the class-imbalance problem that is usually caused by an overwhelmingly big number of background detections. 
Moreover, the low number of duplicates (\autoref{fig:token_distribution}) further demonstrates the viability of detecting objects without post-processing.


%

\subsection{Effect of Attention}
\begin{figure*}[t]
\centering
\subfigure{
\includegraphics[width=.213\linewidth]{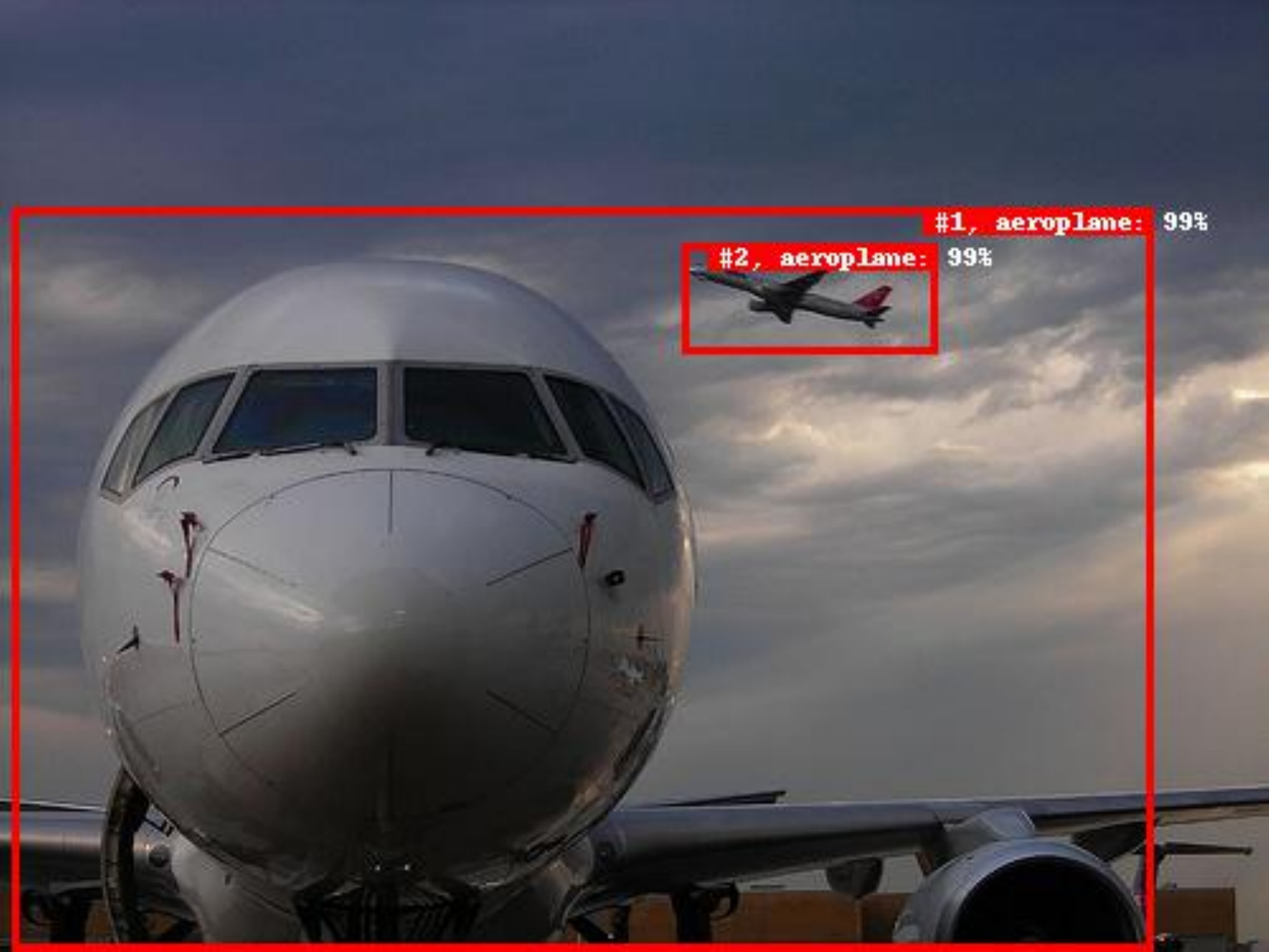}
}
\subfigure{
\includegraphics[width=.213\linewidth]{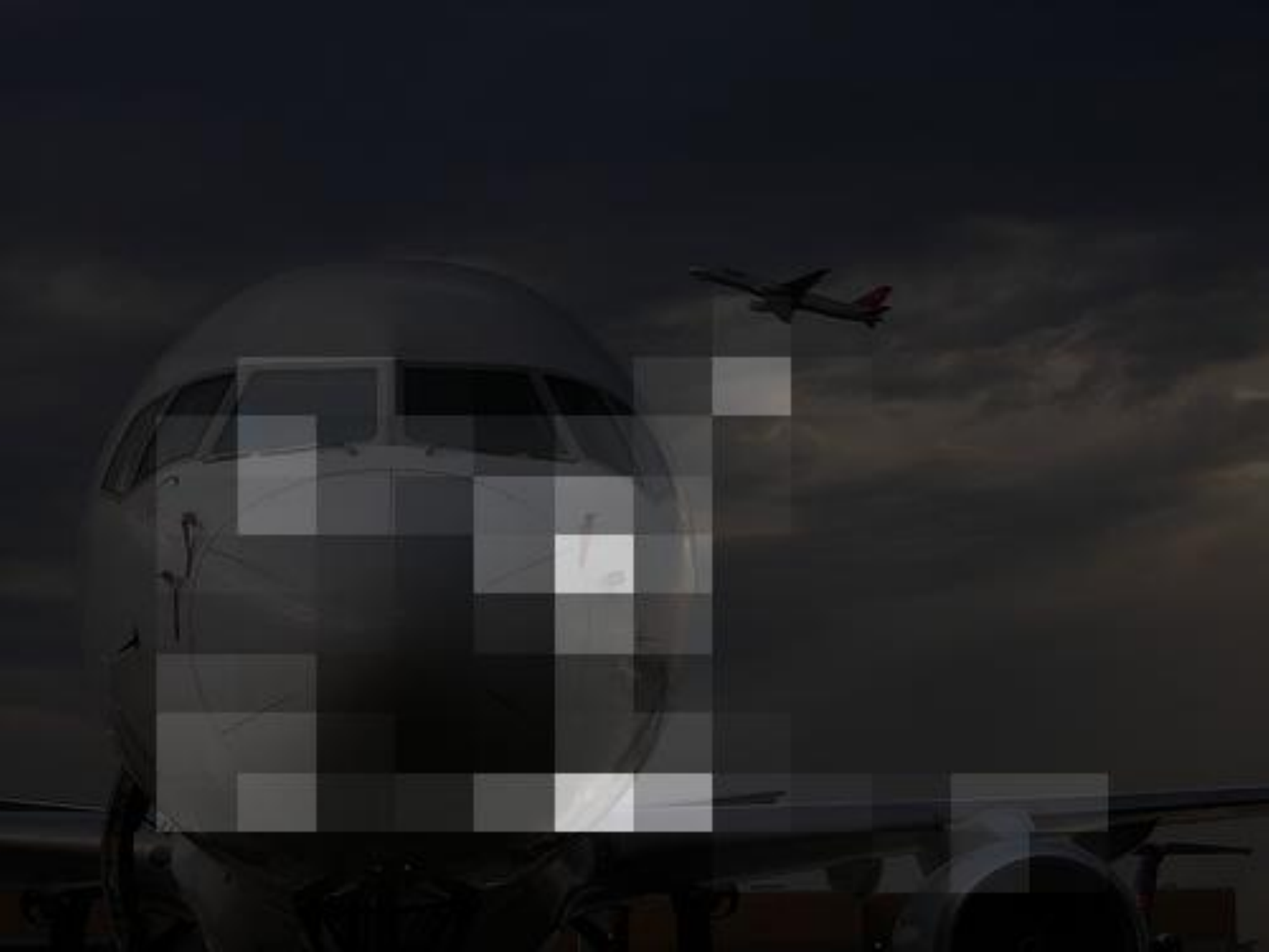}
}
\subfigure{
\includegraphics[width=.213\linewidth]{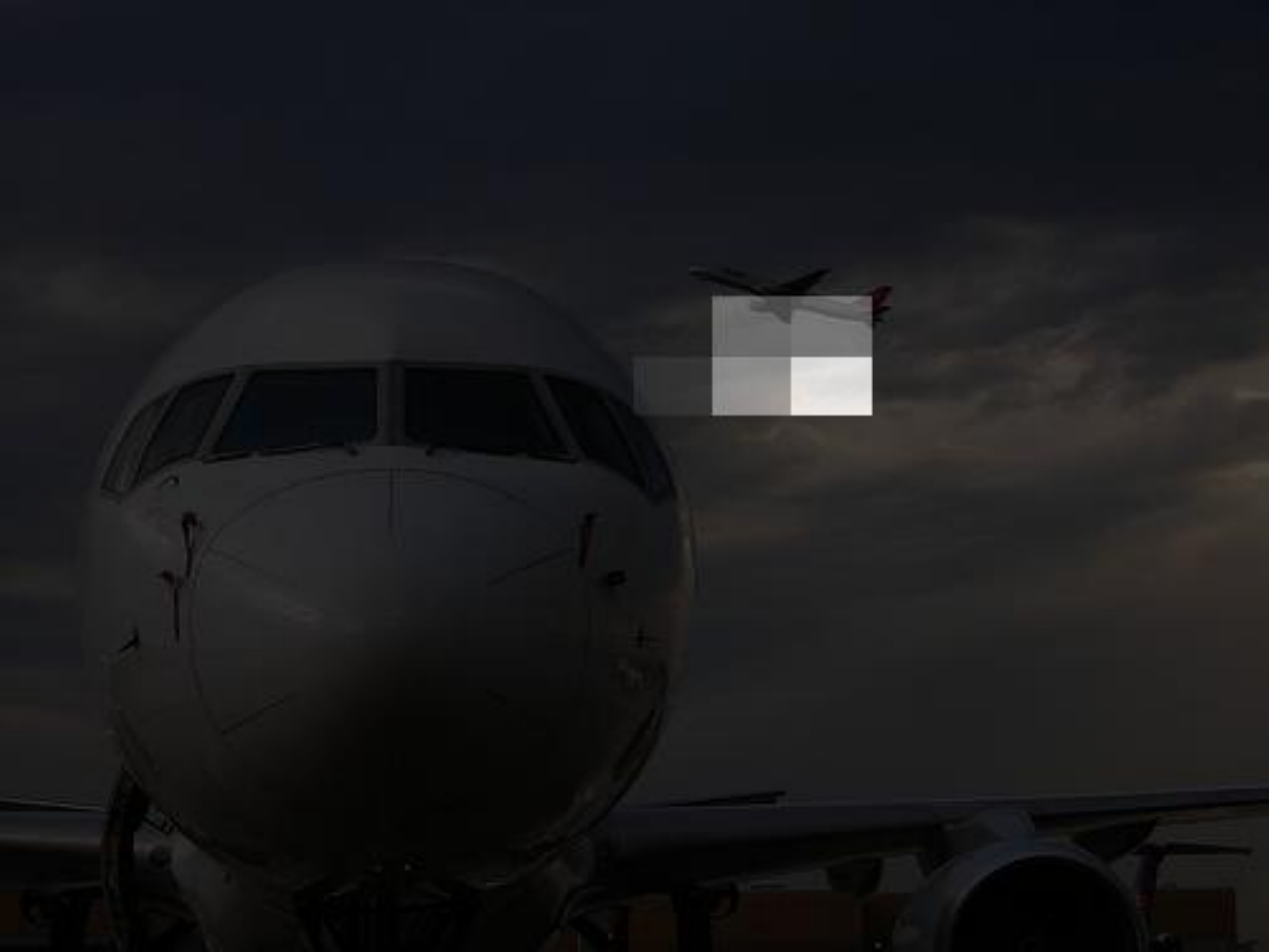}
}

\subfigure{
\includegraphics[width=.155\linewidth]{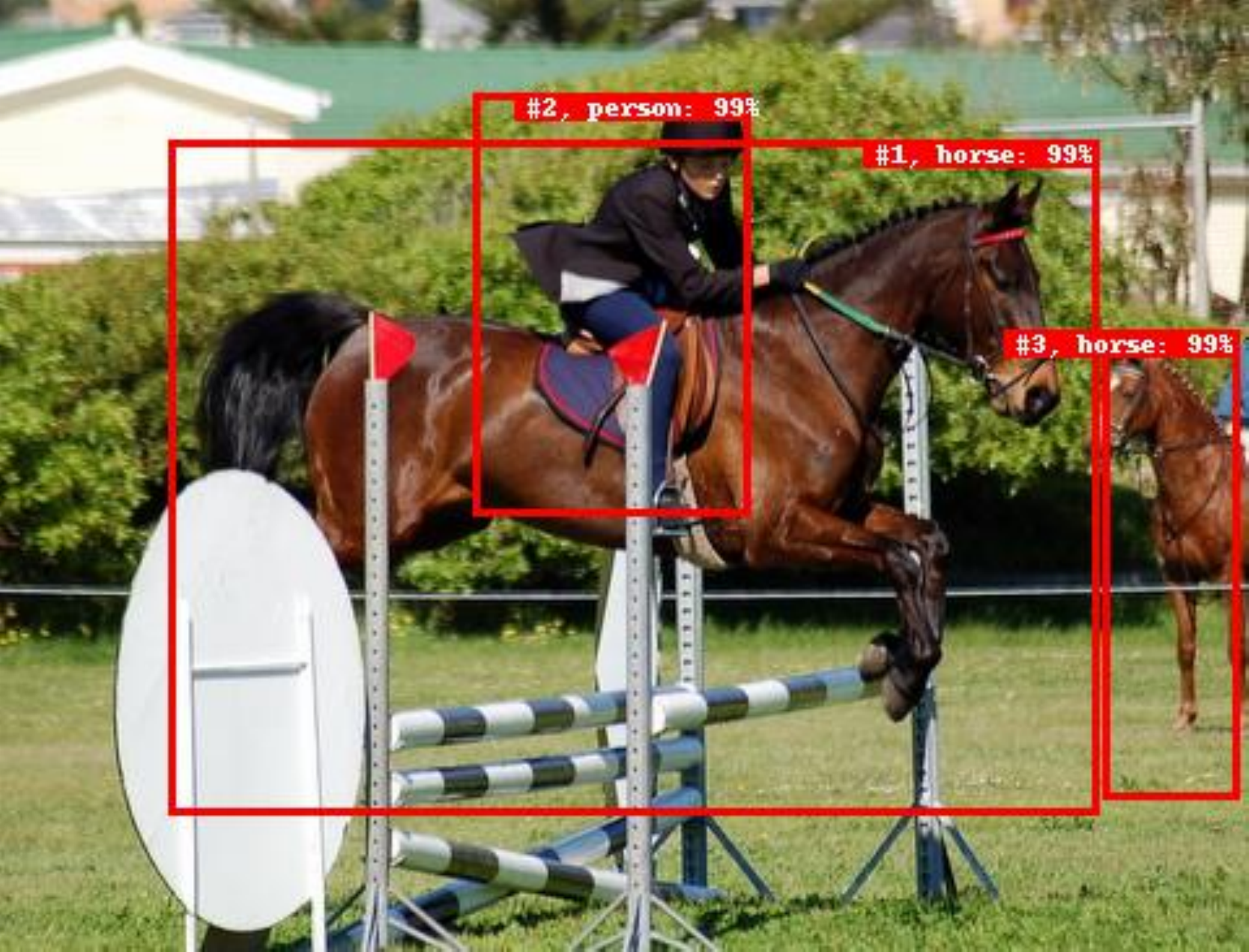}
}
\subfigure{
\includegraphics[width=.155\linewidth]{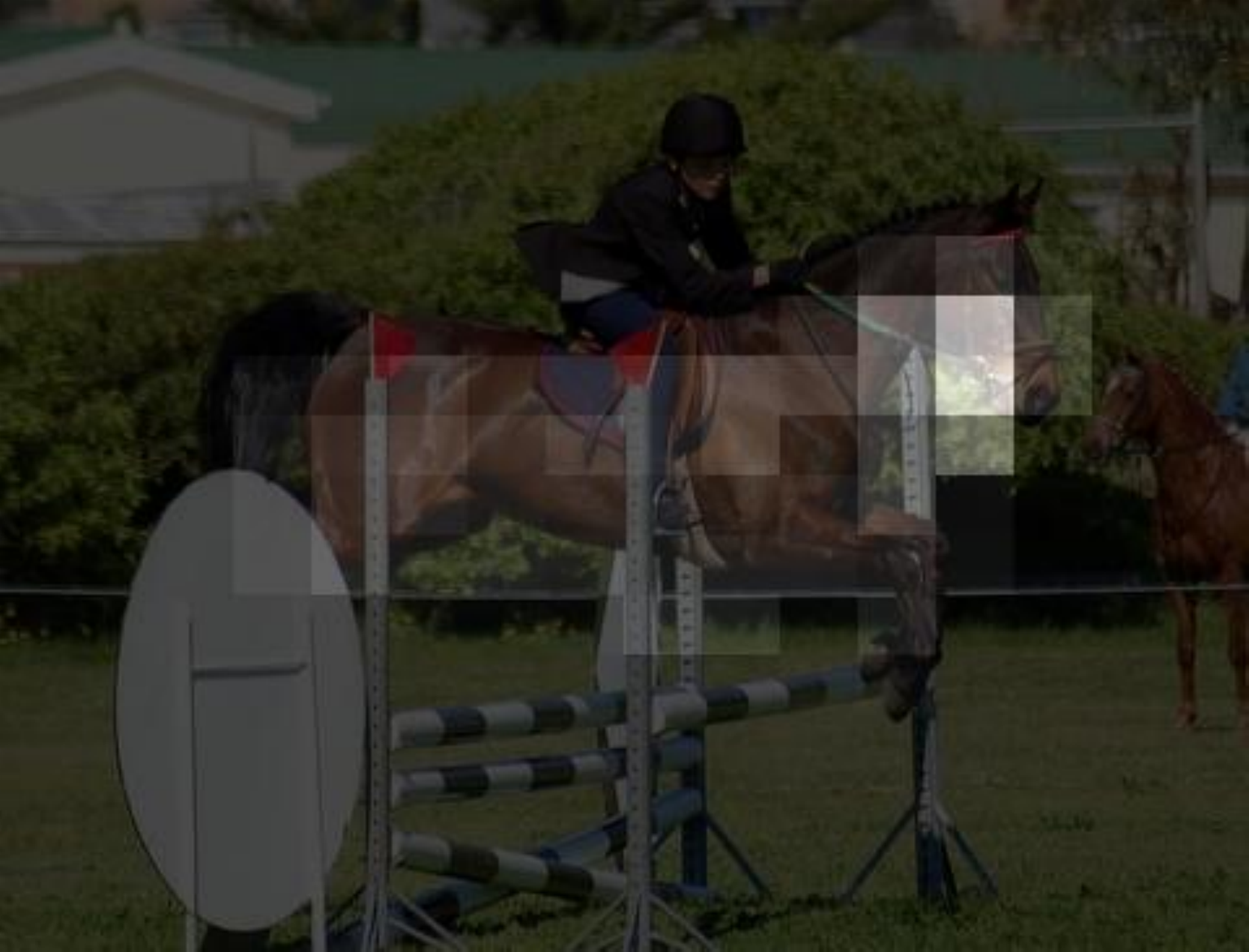}
}
\subfigure{
\includegraphics[width=.155\linewidth]{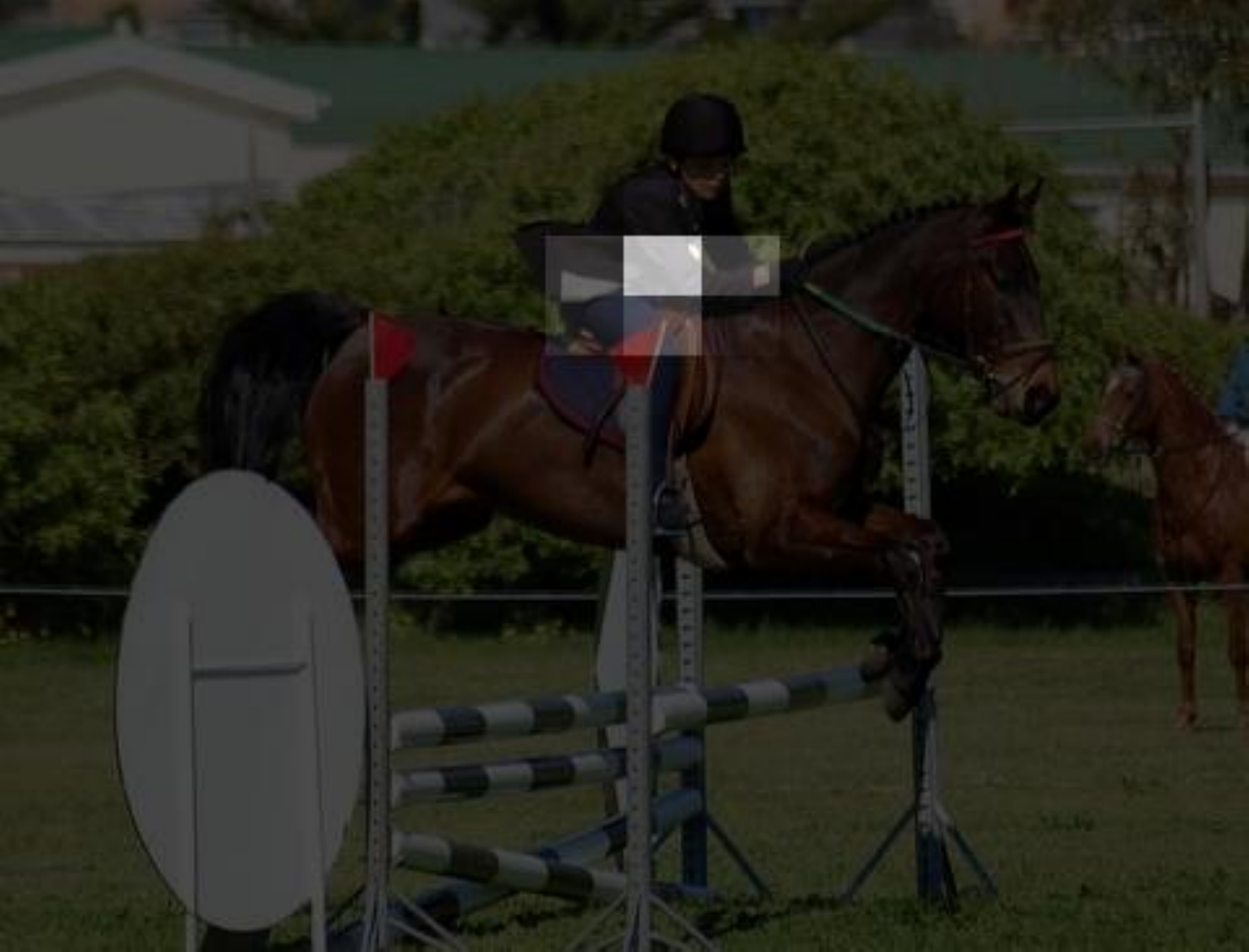}
}
\subfigure{
\includegraphics[width=.155\linewidth]{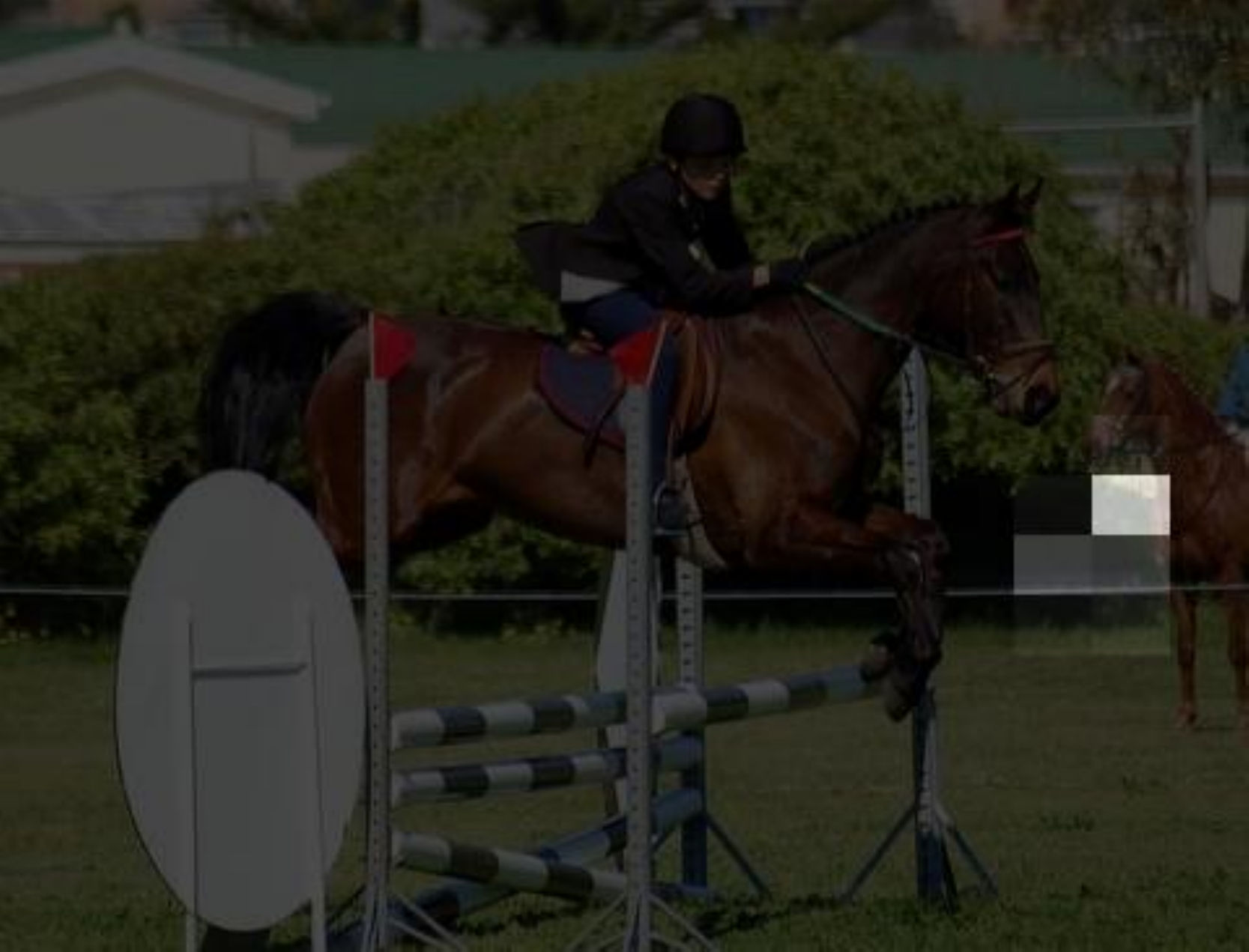}
}

\subfigure{
\includegraphics[width=.155\linewidth]{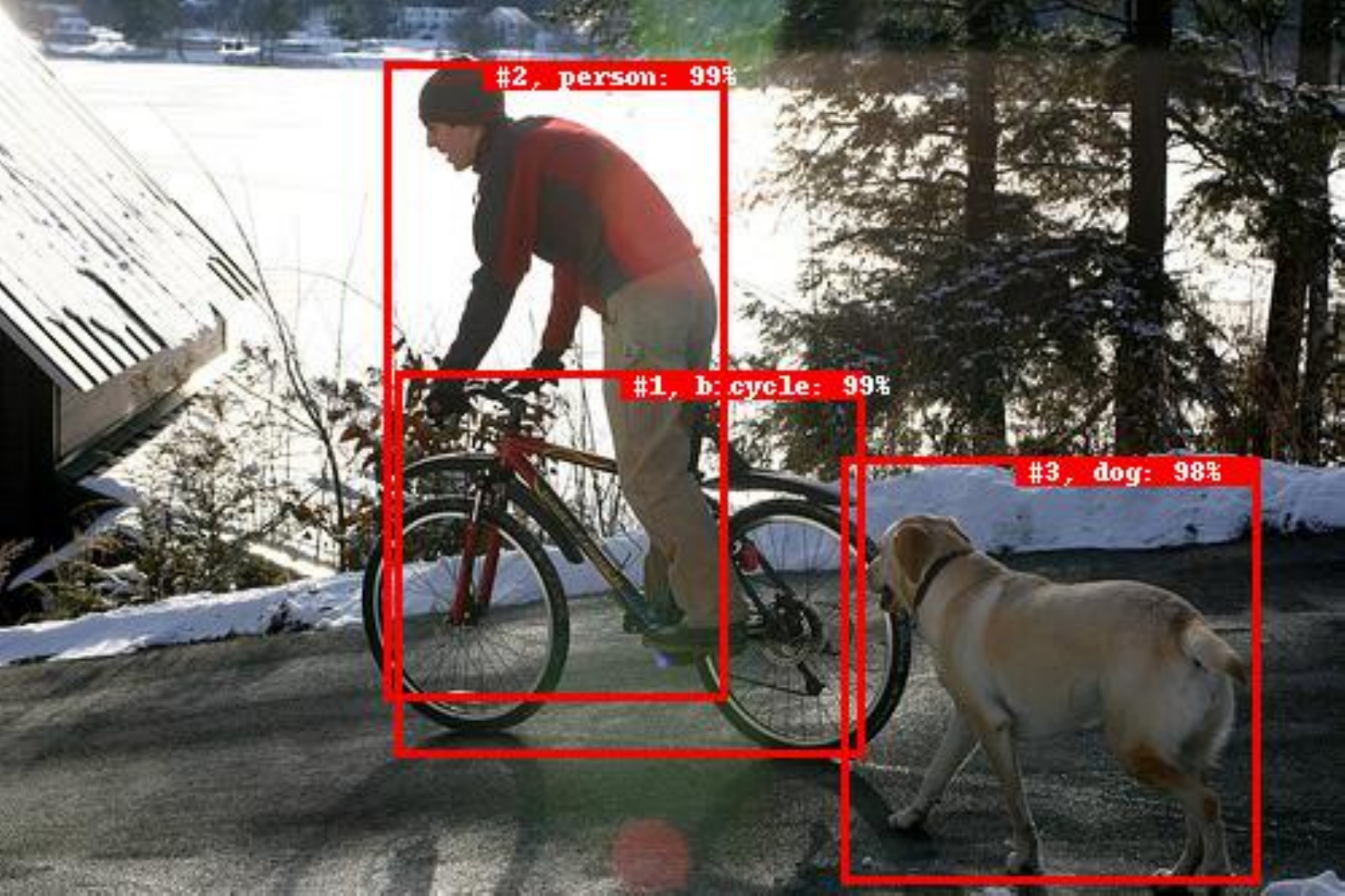}
}
\subfigure{
\includegraphics[width=.155\linewidth]{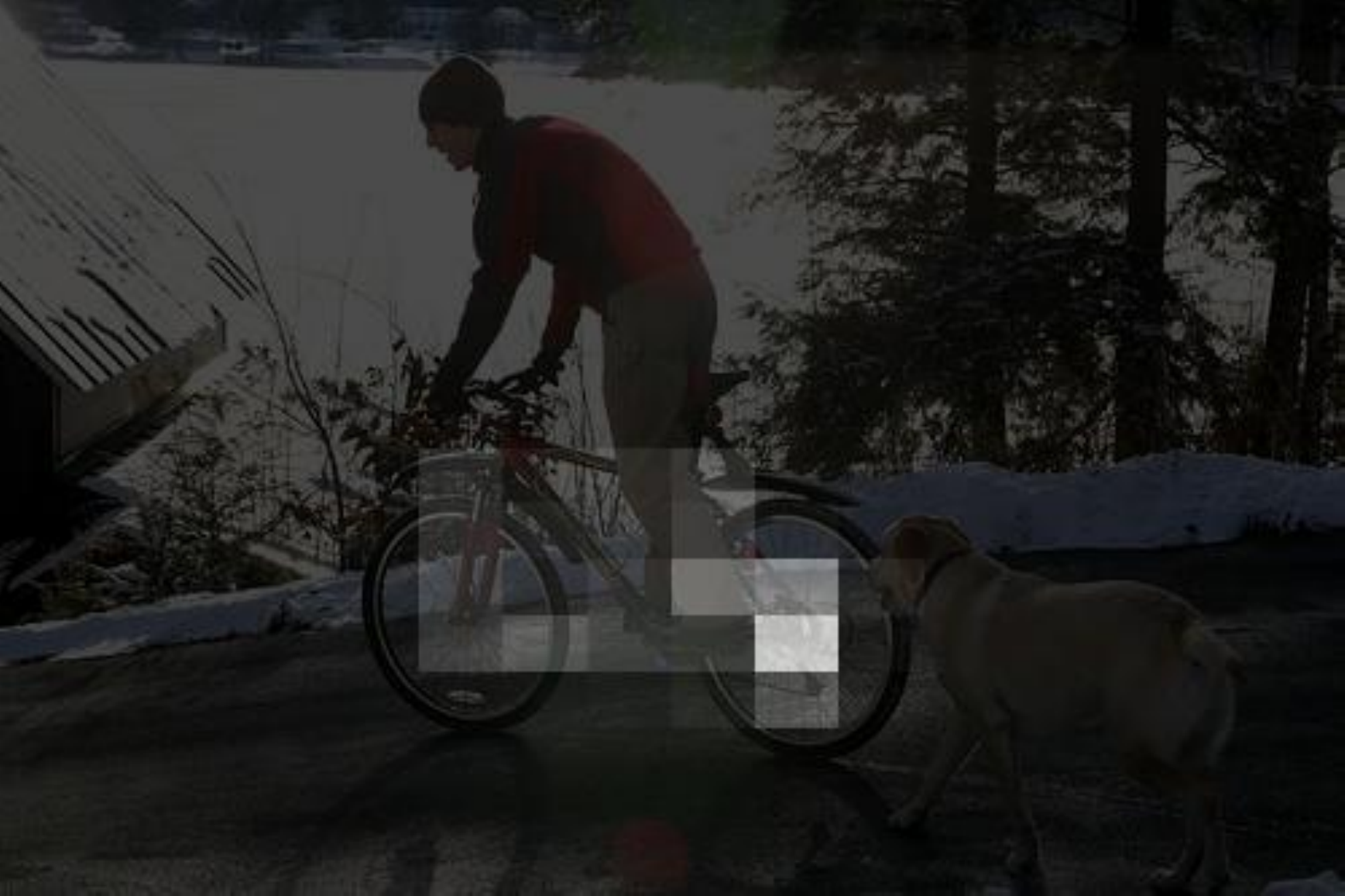}
}
\subfigure{
\includegraphics[width=.155\linewidth]{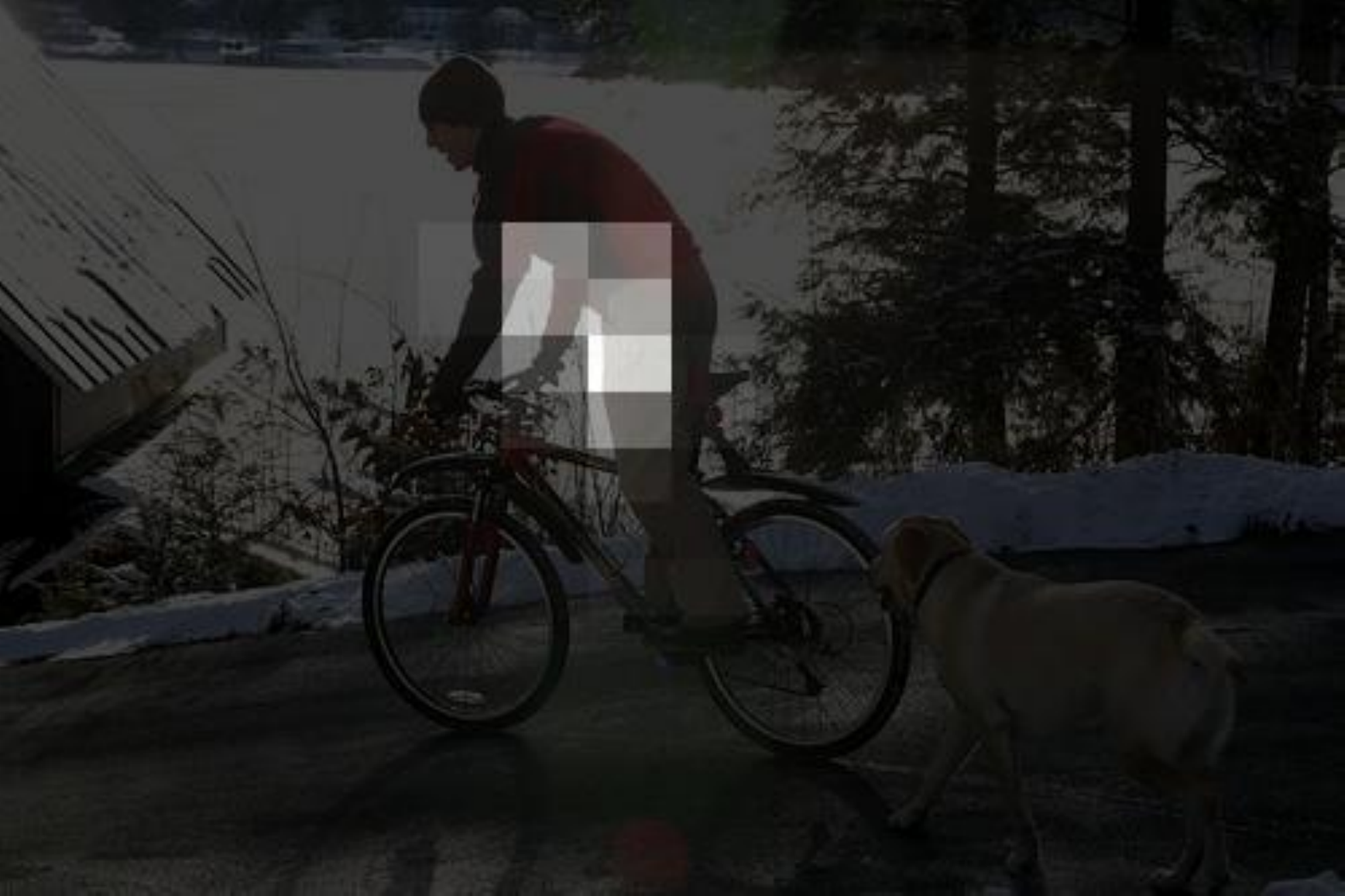}
}
\subfigure{
\includegraphics[width=.155\linewidth]{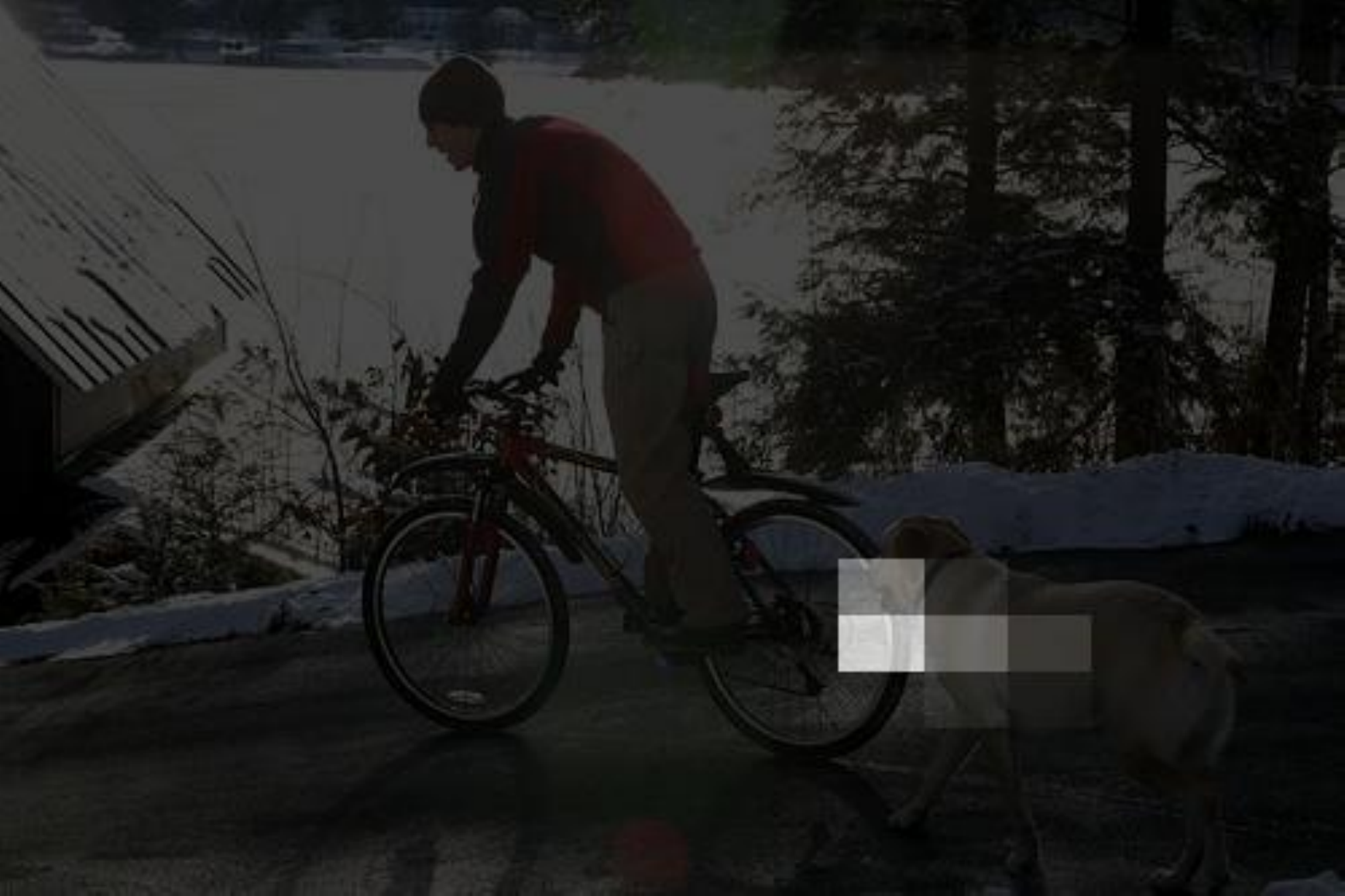}
}

\subfigure{
\includegraphics[width=.155\linewidth]{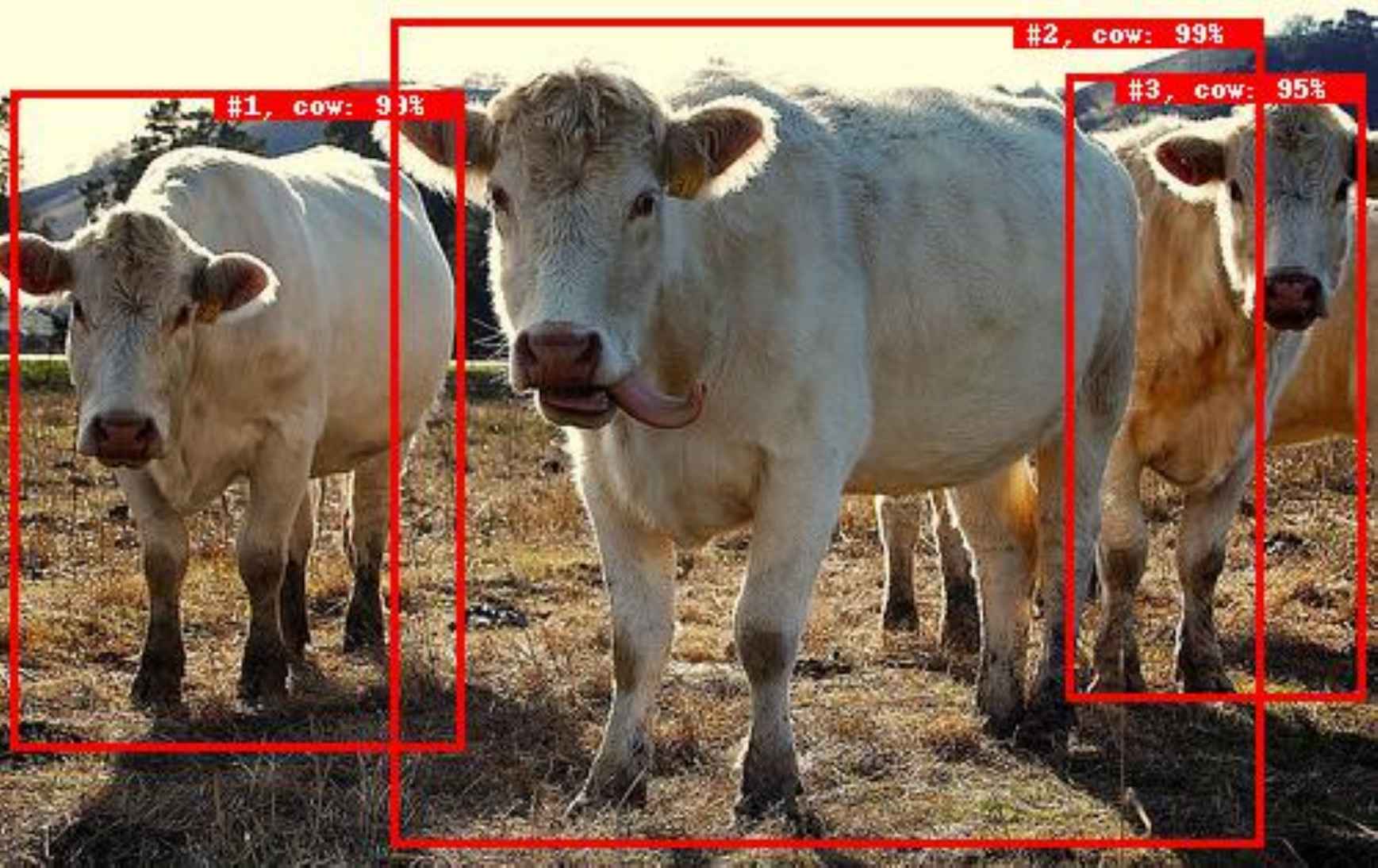}
}
\subfigure{
\includegraphics[width=.155\linewidth]{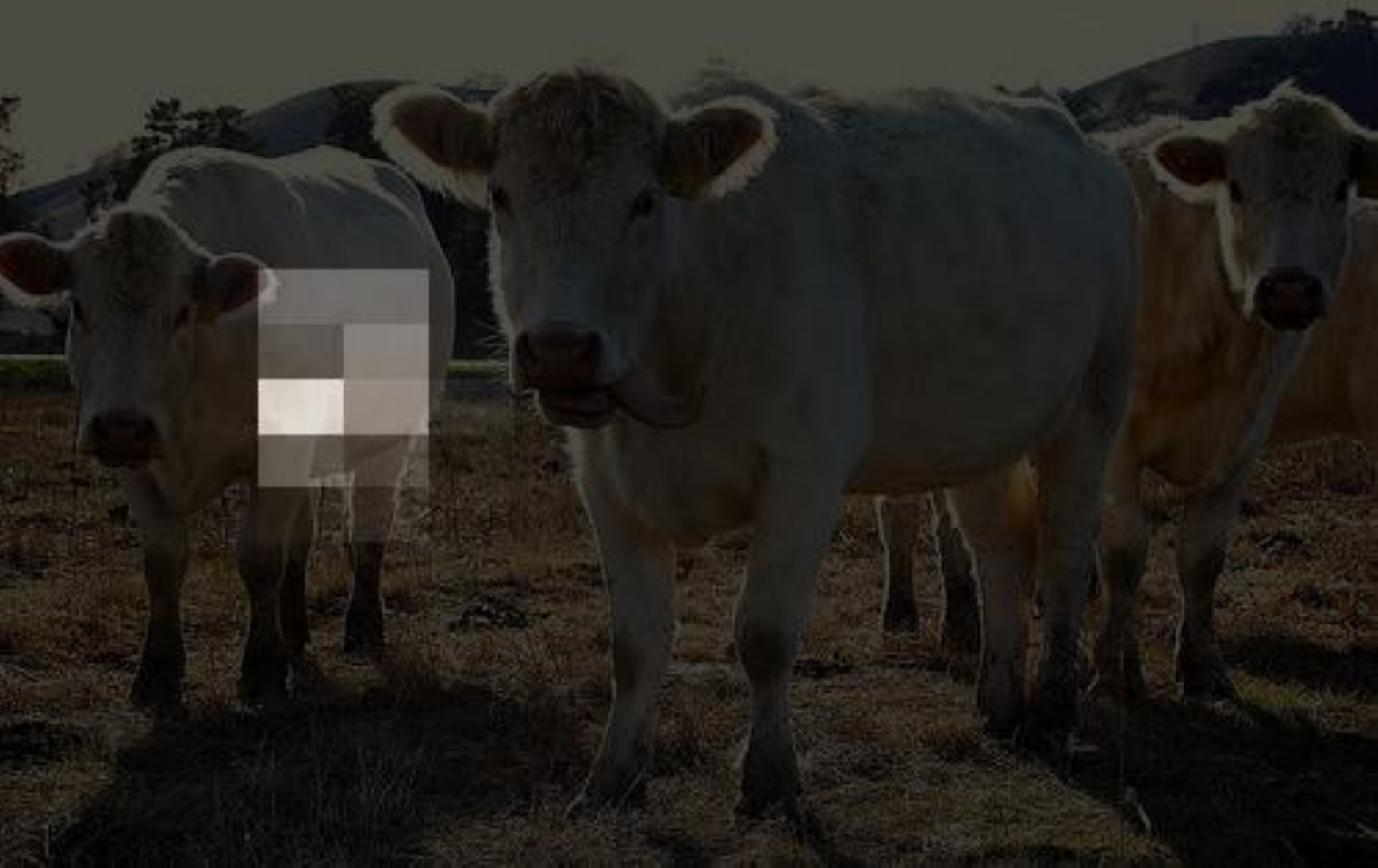}
}
\subfigure{
\includegraphics[width=.155\linewidth]{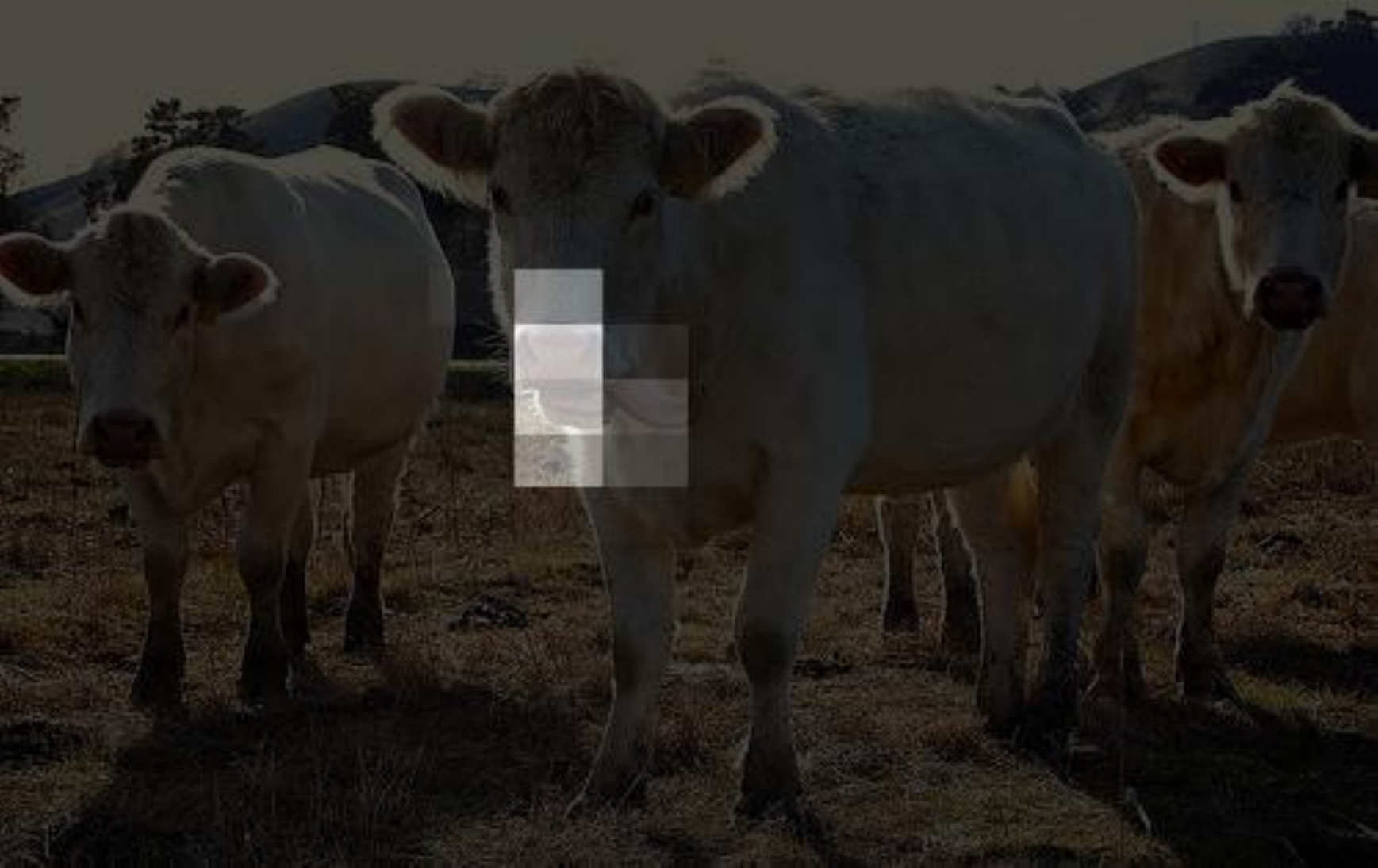}
}
\subfigure{
\includegraphics[width=.155\linewidth]{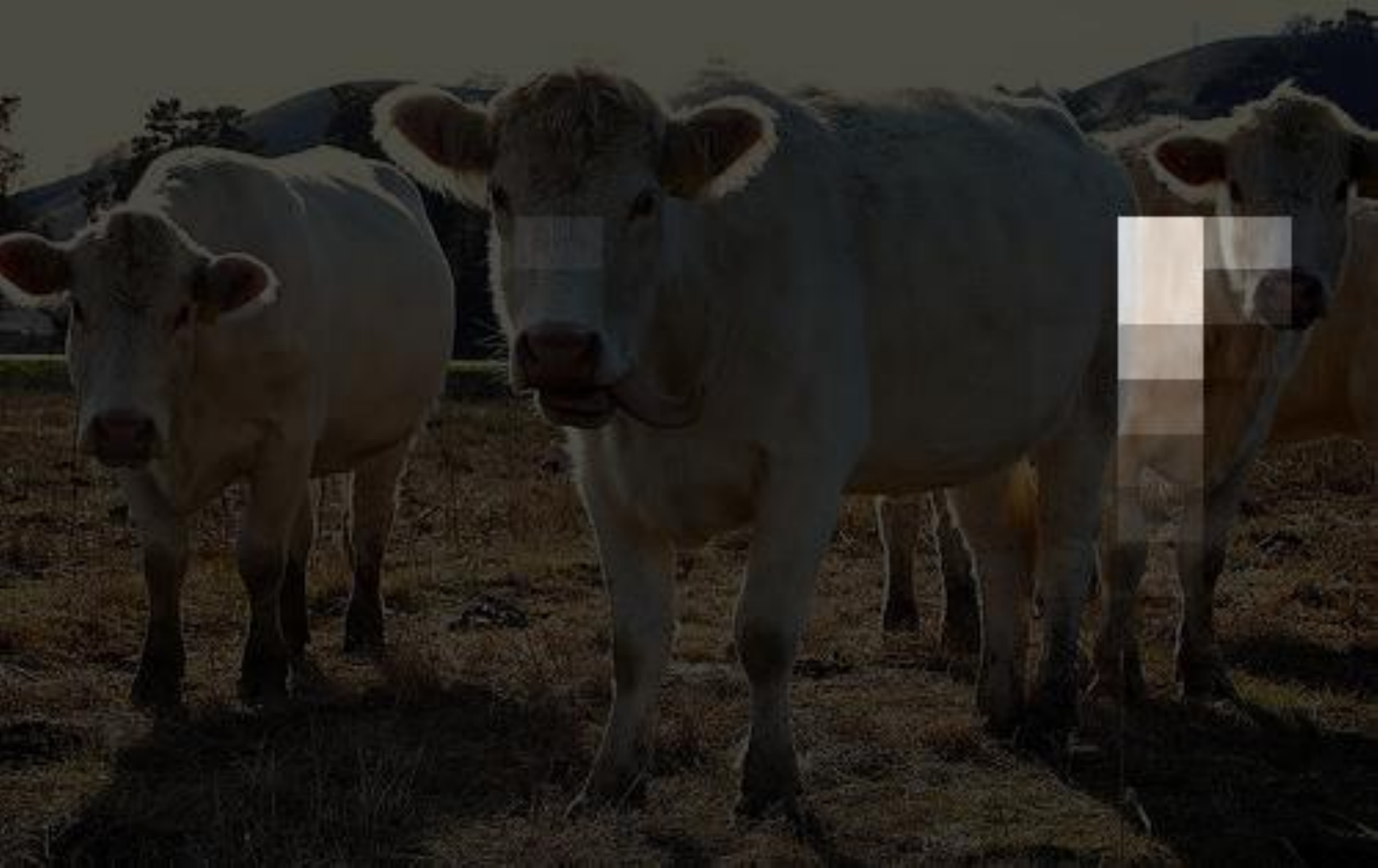}
}

\subfigure{
\includegraphics[width=.12\linewidth]{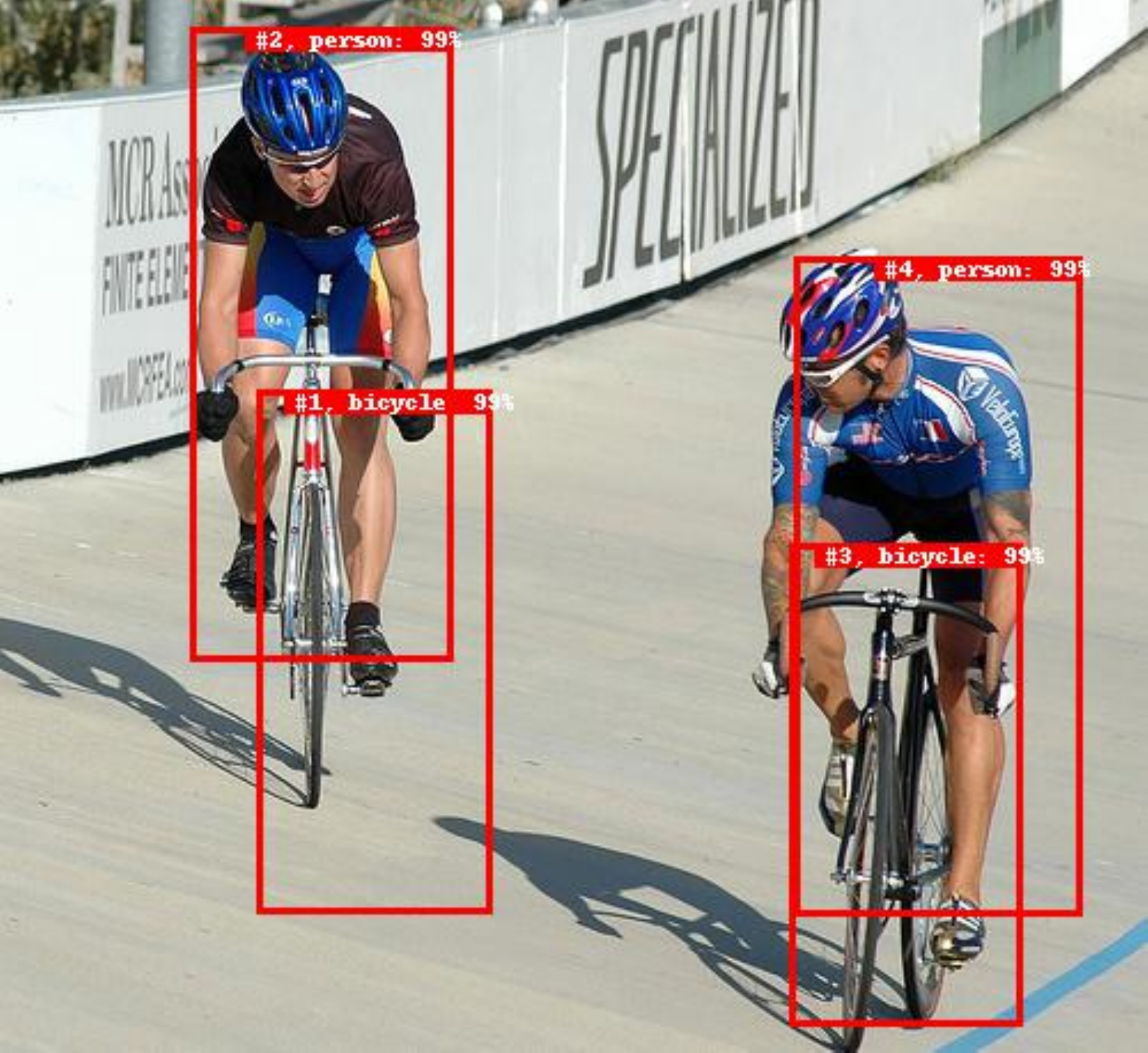}
}
\subfigure{
\includegraphics[width=.12\linewidth]{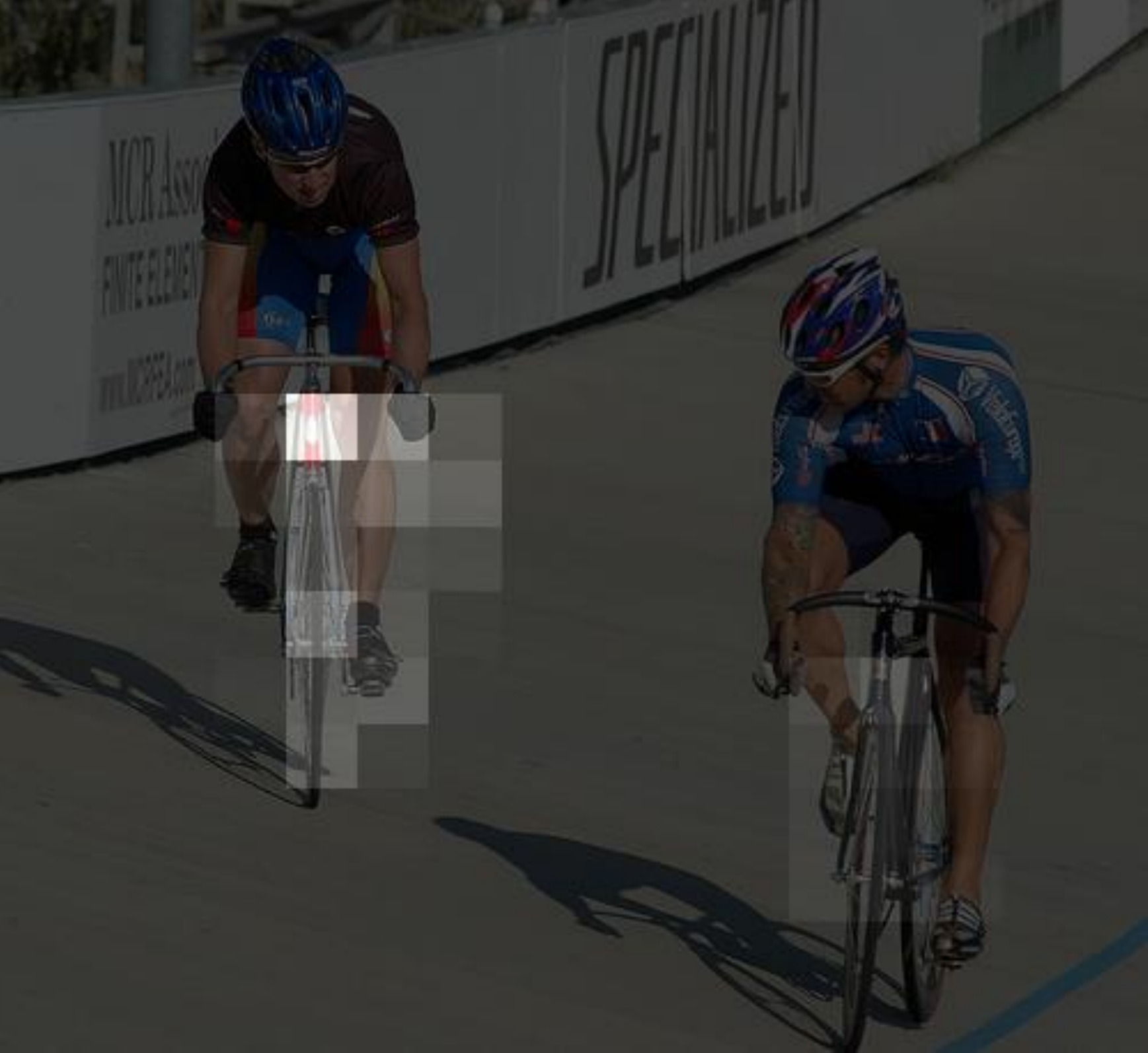}
}
\subfigure{
\includegraphics[width=.12\linewidth]{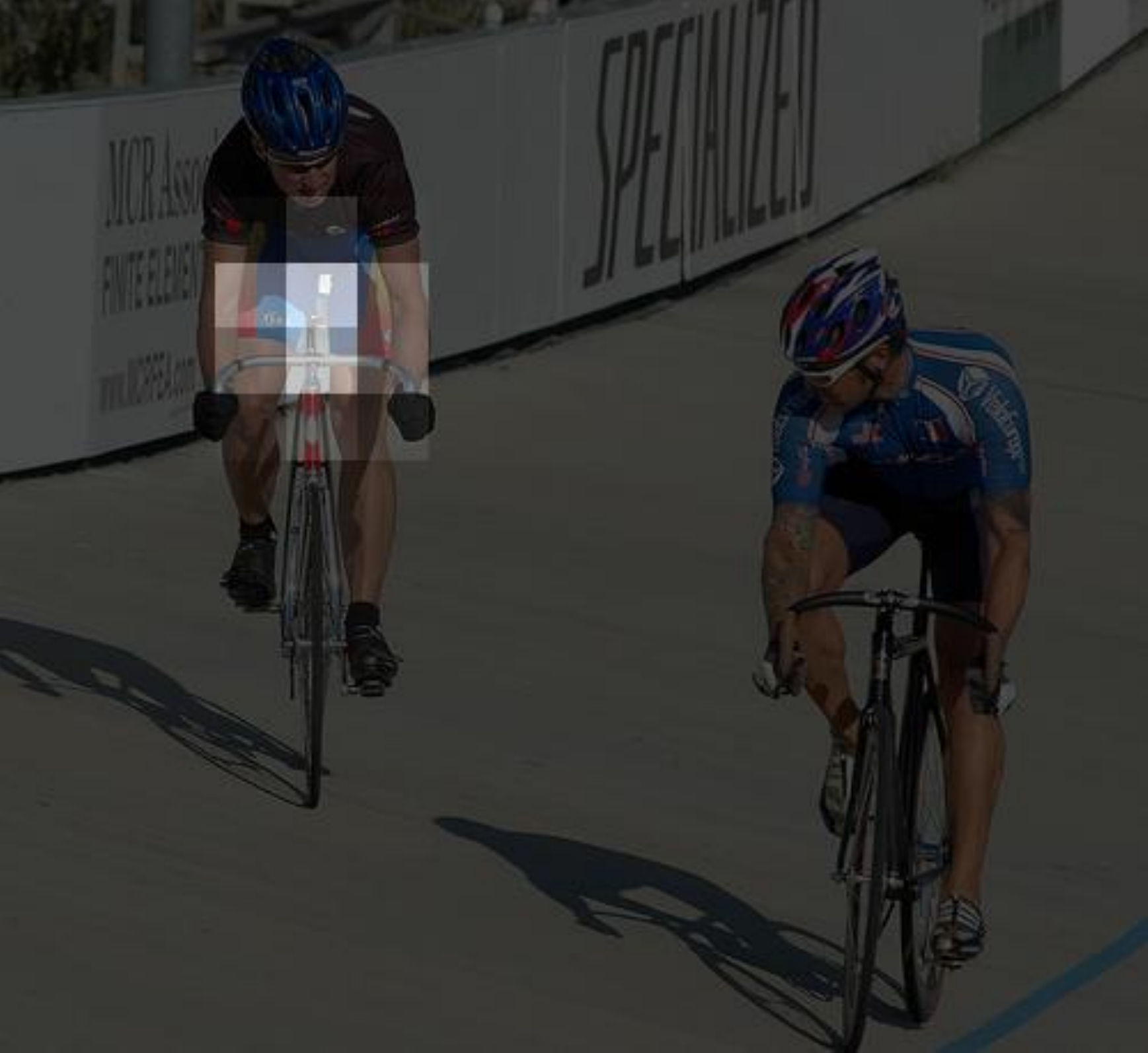}
}
\subfigure{
\includegraphics[width=.12\linewidth]{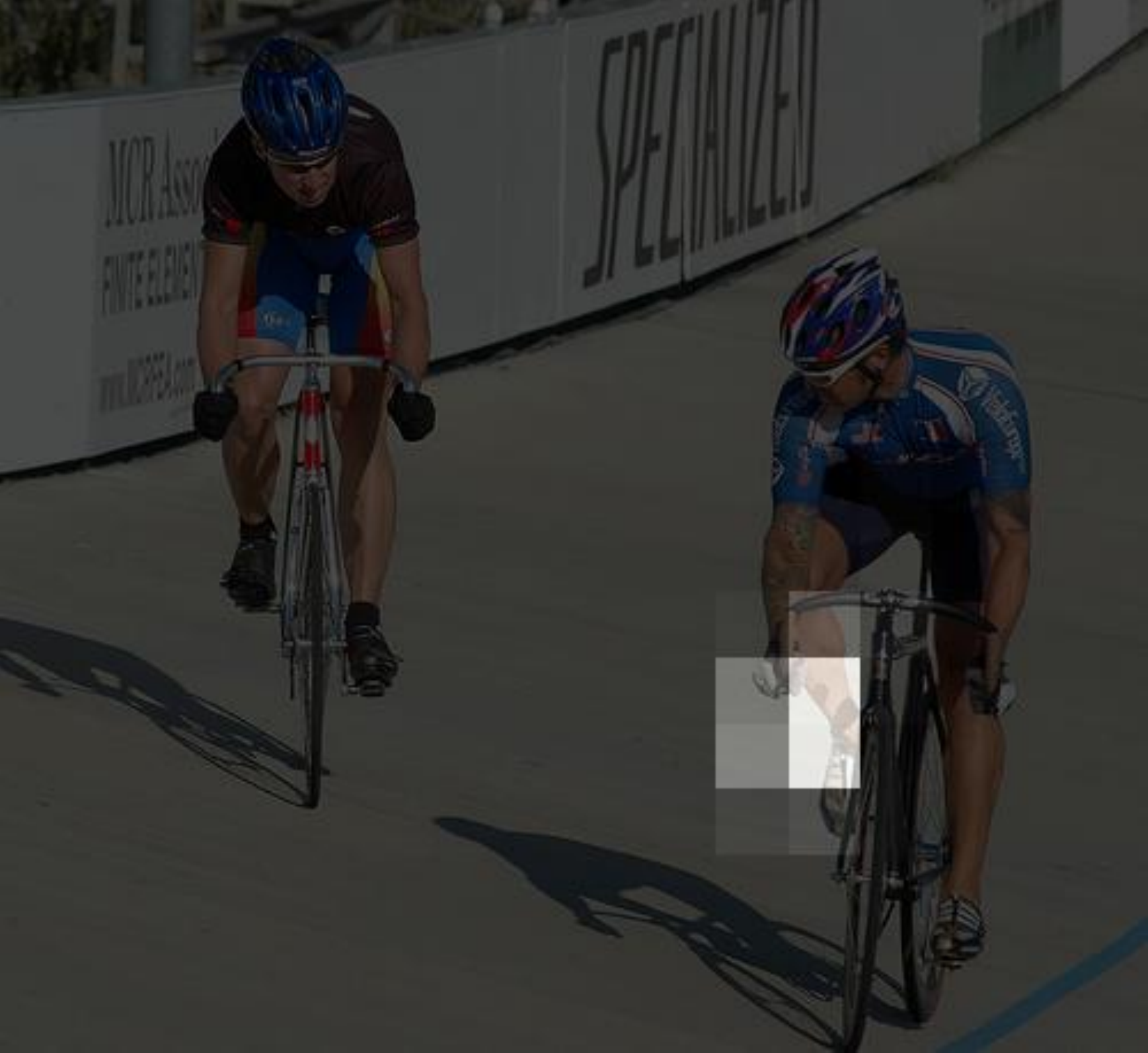}
}
\subfigure{
\includegraphics[width=.12\linewidth]{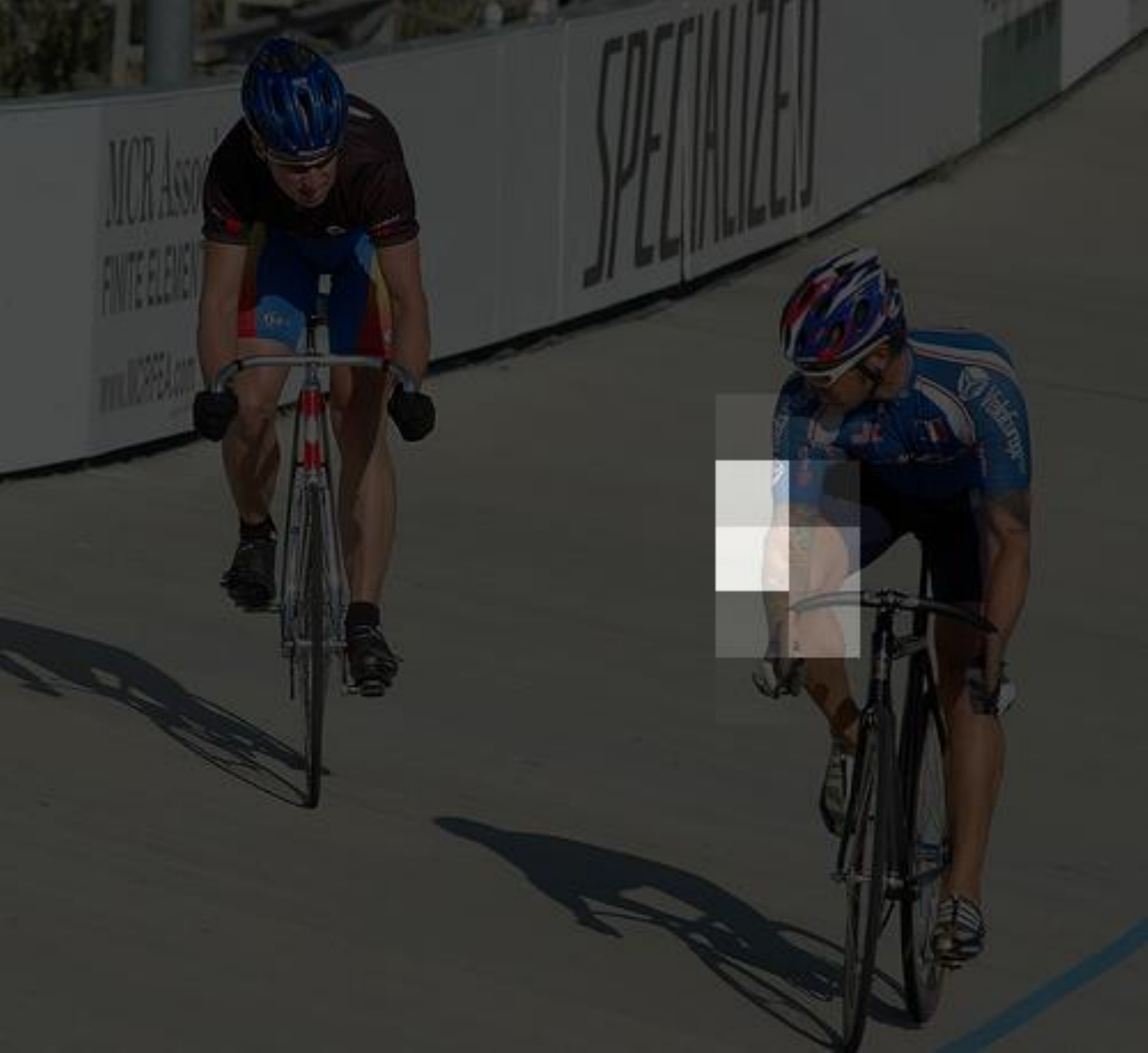}
}

\caption{Example object detection results and attention maps produced by Detective on \texttt{test07}. Left-most images illustrate the detected objects. These correspond to the raw output of Detective after discarding the EoS token and background detections. Subsequent images are the resulting attention maps at each iteration of the ConvLSTM.}
\label{fig:attention}
\end{figure*}

As covered by the ablation study, attention plays an important role in Detective by adding spatial awareness. We qualitatively demonstrate this further by visualizing the attention maps at each iteration on various images as shown in \autoref{fig:attention}. The visualizations illustrate how the attention mechanism selectively shifts the focus at each iteration from one object to the other. In doing so, it allows the network to compute meaningful object representations at each iteration that serve to predict the class and the bounding box offsets of the object. \looseness=-1

\subsection{Potential of the proposed approach}

Although sparse object detection is still challenging, Detective shows promising results as it can be seen in  \autoref{fig:attention}. 
To demonstrate how Detective would behave under a \textit{dense} regime, we allow the network in the following experiment to make an extra number of predictions and set the total number of ConvLSTM iterations during training to $ m = n + 5 $, where $ n $ is the number of ground truth objects. 
During training, the Hungarian algorithm matches the $ n $ targets to $ n $ predictions from the first $ m - 1 $ predictions and the last prediction is matched to the EoS token. 
The loss computation follows the same Equations~\ref{eq:loss},~\ref{eq:loss_cls} and~\ref{eq:loss_loc}, and the predictions that were not matched are simply ignored when computing the loss. 
As a result, the model is not penalized when producing duplicate detections. This alleviates  the constraint of predicting the exact objects in the image, but at the expense of achieving fully end-to-end object detection. At inference, the model produces a higher number of duplicates which need to be filtered using NMS. 
After applying NMS, the model achieves an mAP of $64.1$\%, \ie, $12$\% higher than the mAP achieved by Detective. 
Even though this shows that making dense predictions still outperforms sparse object detection, it attests for the potential of Detective as it significantly gains in performance while being trained to only make $ 5 $ extra predictions. 

\section{Conclusion}
In this work, we proposed Detective -- a novel architecture that tackles the object detection task without requiring any post-processing. 
Detective comprises a ConvLSTM coupled with an attention mechanism, that sequentially detects the object instances in an image and determines when to stop iterating by emitting an EoS token. 
Our network achieves promising results on the popular PASCAL VOC dataset for object detection, showing an improvement of over $11$\% in comparison to a baseline LSTM model.
Finally, by visualizing the generated attention maps, we show how our model selectively focuses on each object instance, which demonstrates how Detective is able to reason at the instance level.\looseness=-1

\bibliographystyle{IEEEtran}
\bibliography{paper}

\end{document}